\documentclass[9pt, onecolumn]{article}

\usepackage[margin=1in]{geometry}  % Adjust margins

\usepackage{cite}
\usepackage{ulem}
\usepackage{xcolor}
\usepackage{acronym}
\usepackage{caption}
\usepackage{leftidx}
\usepackage{listings}
\usepackage{graphicx}
\usepackage{textcomp}
\usepackage{multirow}
\usepackage{booktabs}
\usepackage{colortbl}
\usepackage{hyperref}
\usepackage{csquotes}
\usepackage{rotating}
\usepackage{subcaption}
\usepackage{algorithmic}
\usepackage{amsmath,amssymb,amsfonts}

\acrodef{VSLAM}{Visual SLAM}
\acrodef{FoV}{Field of View}
\acrodef{CV}{Computer Vision}
\acrodef{BA}{Bundle Adjustment}
\acrodef{FPS}{Frames per Second}
\acrodef{STD}{Standard Deviation}
\acrodef{GNN}{Graph Neural Network}
\acrodef{LSD}{Line Segment Detector}
\acrodef{ROS}{Robot Operating System}
\acrodef{RGB-D}{Red Green Blue-Depth}
\acrodef{RMSE}{Root Mean Square Error}
\acrodef{ATE}{Absolute Trajectory Error}
\acrodef{IMU}{Inertial Measurement Unit}
\acrodef{RANSAC}{RANdom SAmple Consensus}
\acrodef{CNN}{Convolutional Neural Network}
\acrodef{LiDAR}{Light Detection And Ranging}
\acrodef{ORB}{Oriented FAST and Rotated BRIEF}
\acrodef{API}{Application Programming Interface}
\acrodef{SLAM}{Simultaneous Localization and Mapping}

\newcommand{\orb}{ORB-SLAM 3.0}

\newcommand{\sgraphs}{\textit{S-Graphs}}
\newcommand{\vgraphs}{\textit{vS-Graphs}}
\definecolor{red}{HTML}{fd8f8f}
\definecolor{greend}{HTML}{57e377}
\definecolor{greenl}{HTML}{b8fb8a}
\definecolor{lyellow}{HTML}{fefdb4}
\definecolor{orange}{HTML}{ffd5ab}
\definecolor{grayl}{HTML}{d5d5d5}

\colorlet{red}{red!50}
\colorlet{yellow}{yellow!50}
\colorlet{greenl}{greenl!50}
\colorlet{greend}{greend!50}
\colorlet{orange}{orange!50}
\colorlet{grayl}{grayl!25}
\colorlet{lyellow}{lyellow!50}

\title{\LARGE \bf vS-Graphs: Tightly Coupling Visual SLAM and 3D Scene Graphs Exploiting Hierarchical Scene Understanding}

\author{
    Ali Tourani$^{1}$, Saad Ejaz$^{1}$, Hriday Bavle$^{1}$, Miguel Fernandez-Cortizas$^{1}$, \\ David Morilla-Cabello$^{2}$, Jose Luis Sanchez-Lopez$^{1}$, and Holger Voos$^{1}$
    \thanks{$^{1}$Authors are with the Automation and Robotics Research Group, Interdisciplinary Centre for Security, Reliability, and Trust (SnT), University of Luxembourg, Luxembourg. Ali Tourani is also associated with the Institute for Advanced Studies (IAS), University of Luxembourg, Luxembourg. Holger Voos is also associated with the Faculty of Science, Technology, and Medicine, University of Luxembourg, Luxembourg. \tt{\small{\{ali.tourani, saad.ejaz, hriday.bavle, miguel.fernandez, joseluis.sanchezlopez, holger.voos\}}@uni.lu}}
    \thanks{$^{2}$Author is with the Instituto de Investigación en Ingeniería de Aragón (I3A), Universidad de Zaragoza, Spain. {\tt\small davidmc@unizar.es}}
    \thanks{*This research was funded, in whole or part, by the Luxembourg National Research Fund (FNR), DEUS Project, ref. C22/IS/17387634/DEUS. It was also partially funded by the Institute of Advanced Studies (IAS) of the University of Luxembourg through an “Audacity” grant (project TRANSCEND).}
    \thanks{*For the purpose of open access, and in fulfillment of the obligations arising from the grant agreement, the author has applied a Creative Commons Attribution 4.0 International (CC BY 4.0) license to any  Author Accepted Manuscript version arising from this submission.}
}

\begin{document}

\maketitle
\thispagestyle{empty}
\pagestyle{empty}

\begin{abstract}
Current Visual Simultaneous Localization and Mapping (VSLAM) systems often struggle to create maps that are both semantically rich and easy to interpret.
While incorporating semantic scene knowledge helps build richer maps with contextual associations among mapped objects, representing them in structured formats such as scene graphs has not been widely addressed, leading to complex map comprehension and limited scalability.
This paper introduces vS-Graphs, a novel real-time VSLAM framework that integrates vision-based scene understanding with map reconstruction and comprehensible graph-based representation.
The framework infers structural elements (\textit{i.e.,} rooms and floors) from detected building components (\textit{i.e.,} walls and ground surfaces) and incorporates them into optimizable 3D scene graphs.
This solution enhances the reconstructed map's semantic richness, comprehensibility, and localization accuracy.
Extensive experiments on standard benchmarks and real-world datasets demonstrate that vS-Graphs achieves an average of $\mathbf{15.22\%}$ accuracy gain across all tested datasets compared to state-of-the-art VSLAM methods.
Furthermore, the proposed framework achieves environment-driven semantic entity detection accuracy comparable to that of precise LiDAR-based frameworks, using only visual features.

% Web page
The code is publicly available at \url{https://github.com/snt-arg/visual_sgraphs} and is actively being improved.
Moreover, a web page with additional media and evaluation results is available at \href{https://snt-arg.github.io/vsgraphs-results/}{https://snt-arg.github.io/vsgraphs-results/}.
\end{abstract}
\section{Introduction}
\label{sec_intro}

% Intro and VSLAM
Robust environment understanding, a core foundation of robots' situational awareness \cite{slamtosa} in the context of \ac{SLAM}, relies heavily on sensor quality and modality.
While diverse sensors, \textit{e.g.,} \ac{LiDAR} and cameras, have been employed in \ac{SLAM}, vision sensors offer a cost-effective solution for rich map reconstruction, forming a distinct category titled \ac{VSLAM} \cite{pu2023visual}.
Among vision sensors, RGB-D cameras provide complementary visual and depth cues, overcoming monocular limitations by producing dense point clouds that capture fine spatial details, enabling precise detection, localization, and mapping of environmental elements \cite{cai2024comprehensive}.
To further enhance \ac{VSLAM} performance, visual perception techniques are integrated, ranging from semantic scene understanding to the incorporation of artificial landmarks such as fiducial markers \cite{semuco, vsgraphs1}.

\begin{figure}[!t]
    \centering
    \includegraphics[width=.65\textwidth]{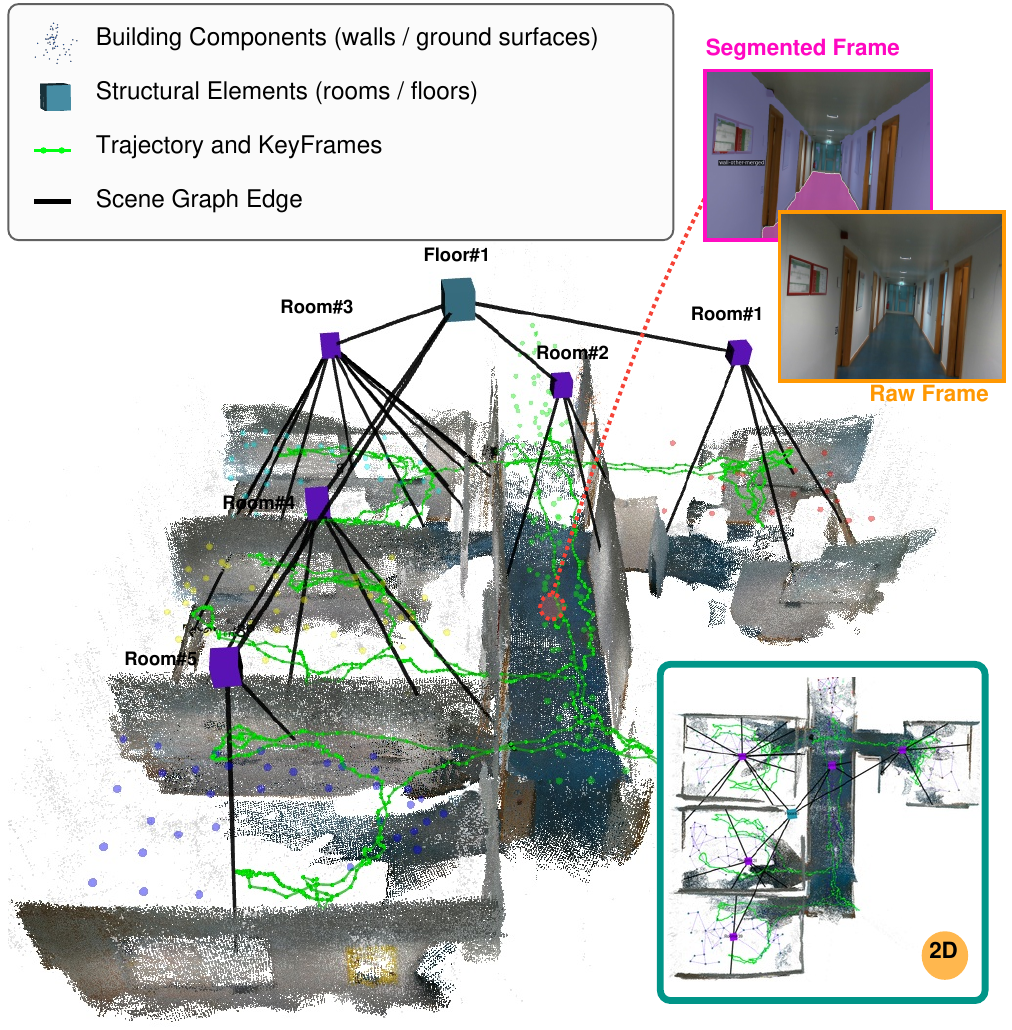}
    \caption{A reconstructed map with its corresponding optimizable 3D scene graph enriched with environment-driven semantic entities, generated by vS-Graphs (SMapper \textit{MR03} sequence).}
    \label{fig_overall}
\end{figure}

% Scene Graph
Beyond enriching maps with visual and depth information, various methodologies aim to organize this data into interpretable and structured representations.
Among them, \textit{scene graphs} represent environments as hierarchical structures that capture the presence of objects, their attributes, and inter-relationships.
These solutions provide a higher level of abstraction for scene understanding by outlining spatial associations among observed entities \cite{3ddsg}.
While scene graph-driven works like \cite{sgrec3d, scenegraphfusion} focus on tailoring geometric and semantic cues for reliable interpretation, approaches such as Hydra \cite{hydra} and HOV-SG \cite{hovsg} serve as 3D scene graph builders (rather than SLAM systems), representing spatial hierarchies without estimating camera trajectories.
Moreover, HOV-SG requires ground-truth semantics and an entire-environment map to construct the scene graph offline.
In contrast, works like \textit{S-Graphs} \cite{sgraphsp} push the boundaries by directly incorporating scene graphs into the SLAM pipeline, relying on LiDAR odometry with planar surface extraction within a unified optimization system, though without support for visual input.

% Proposed
Inspired by LiDAR-based \textit{S-Graphs} \cite{sgraphsp}, this paper proposes a \ac{VSLAM} framework, \textbf{visual S-Graphs} (vS-Graphs), that tightly couples scene graph construction with VSLAM through joint optimization in a unified factor graph.
vS-Graphs operates in real time and employs both visual and depth data to enhance map reconstruction and camera pose estimation.
It reliably incorporates ``building components'' (\textit{i.e.,} wall and ground surfaces), ``structural elements'' (\textit{i.e.,} rooms and floors), and their spatial associations to produce a more geometry-aware and semantically coherent environment representation.
Unlike LiDAR-based S-Graphs, which rely purely on geometric cues, vS-Graphs exploits the richer semantic and appearance information available in visual data to validate and contextualize structural elements.
This not only strengthens category verification but also establishes a scalable foundation for incorporating additional components in future extensions.
The result is an optimizable, hierarchical 3D scene graph generated online alongside trajectory estimation, pairing robot poses from the underlying SLAM pipeline with recognized semantic entities, as depicted in Fig.~\ref{fig_overall}.
With this, the contributions of the paper are:
\begin{itemize}
    \item A real-time multi-threaded \ac{VSLAM} framework that generates optimizable 3D scene graphs during map reconstruction,
    \item A vision-based method for recognizing and mapping building components (\textit{i.e.,} wall and ground surfaces), enhancing map richness and trajectory estimation,
    \item A mechanism for extracting high-level structural elements from the localized building components for advancing scene understanding; and
    \item Publicly available source code to facilitate reproducibility and further research in the field.
\end{itemize}
\section{Related Works}
\label{sec_related}

% Visual SLAM
Recent advances in computer vision and the development of reliable VSLAM frameworks, such as ORB-SLAM 3.0 \cite{orbslam3}, have enabled more robust localization and mapping systems.
As noted by \cite{tourani2022visual}, depth data is essential for enhancing scene understanding and supporting downstream tasks such as environment modeling.
In this regard, ElasticFusion \cite{elasticfusion} constructs globally consistent surfel maps for photometric tracking, but struggles with scalability and limited integration of environment-level cues.
BAD SLAM \cite{badslam} improves map and trajectory optimization via direct RGB-D \ac{BA}, yet remains sensitive to initialization.
More recent systems, like GS-SLAM \cite{gsslam}, RTG-SLAM \cite{rtgslam}, and SplaTAM \cite{splatam}, incorporate 3D Gaussian representations to enhance tracking and map reconstruction.
Nonetheless, they overlook the scene's semantic context.

% Semantic
Building on these advances, RGB-D VSLAM frameworks have begun integrating semantic awareness into their mapping pipelines.
Tools like Voxblox++ \cite{voxbloxp} augment online scanning with volumetric, object-centric mapping to improve recognition of scene elements.
NICE-SLAM \cite{niceslam} and NIS-SLAM \cite{nisslam} combine neural implicit and hierarchical representations with pre-trained geometric priors to improve dense scene reconstruction at the cost of heavy computation and increased localization errors.
Peng \textit{et al.} \cite{peng2025high} integrate Asymmetric Non-local Neural (ANN)-based semantic segmentation with weighted $K$-means depth clustering for static dense point cloud mapping, but their approach is not real-time, requiring an average of $100.8~\mathrm{ms}$ per frame.
DROID-SLAM \cite{droidslam} uses a fully differentiable pipeline to jointly refine camera poses and dense depth via Gauss–Newton updates driven by optical flow consistency, achieving strong accuracy at the expense of substantial memory and computational overhead.
Methods such as OVD-SLAM \cite{ovdslam} and YDD-SLAM \cite{yddslam} use \acp{CNN} to filter feature points associated with specific semantic objects, thereby refining pose estimation and trajectory accuracy.
Similarly, \cite{baslslam} introduces a binary CNN-based descriptor for robust feature matching and for improving initial pose measurements.
SSF-SLAM \cite{ssfslam} fuses semantic cues with scene-flow geometry via depth–density clustering to build 3D semantic point clouds, enabling improved scene understanding.
Despite their pros, these systems remain limited to object-level reasoning rather than constructing comprehensive scene representations enriched with environment-driven entities.

% Marker-based
Another research direction explores fiducial markers as reliable landmarks, offering an alternative to purely vision-based scene interpretation.
Approaches such as \cite{semuco, vsgraphs1} detect and map environment-driven entities labeled with markers, thereby contributing to a more comprehensive understanding of the environment layout.
Nevertheless, their reliance on pre-placed markers restricts their applicability to controlled environments, limiting flexibility in unprepared or dynamic settings.

% Gap
In contrast to prior semantic VSLAM systems that primarily emphasize object-level cues or dense reconstruction, vS-Graphs embeds scene graph construction directly into the VSLAM pipeline.
By explicitly modeling environment-driven entities and their spatial relationships, it jointly optimizes camera poses and semantic structures, enabling coherent and interpretable indoor scene representations.
\section{Proposed Method}
\label{sec_proposed}

\subsection{System Overview}
\label{sec_proposed_overview}

\begin{figure*}[ht]
    \centering
    \includegraphics[width=\textwidth]{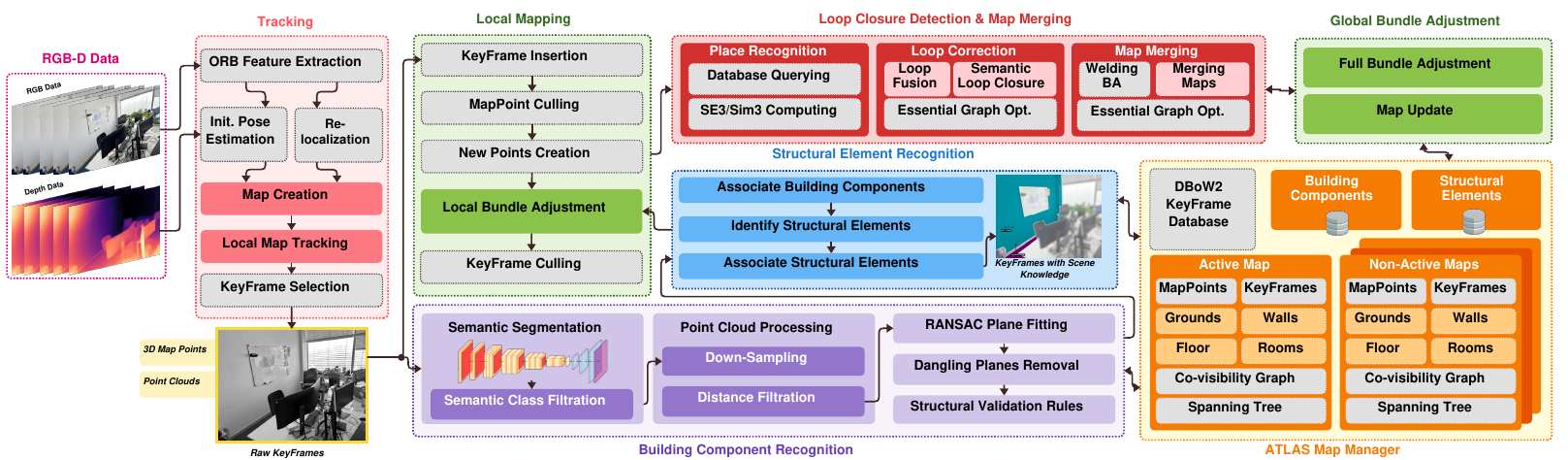}
    \caption{The multi-thread architecture of vS-Graphs. Modules with dashed borders and a light gray background are inherited directly from the baseline (\textit{i.e.,} ORB-SLAM 3.0), while the remaining components are newly added or modified modules.}
    \label{fig_flowchart}
\end{figure*}

% Overal
Building upon ORB-SLAM 3.0, vS-Graphs introduces substantial modifications to its baseline's core modules and adds new threads for robust scene analysis and reconstruction.
The system architecture, shown in Fig.~\ref{fig_flowchart}, details the individual threads, components, and their interconnections.
The current version supports RGB-D input, using depth data primarily to generate surface point clouds, which are then validated to support robust scene understanding.
The core contribution lies in seamlessly integrating two novel threads: \textit{Building Component Recognition} (\S\ref{sec_proposed_building}) and \textit{Structural Element Recognition} (\S\ref{sec_proposed_structural}).
These threads are deeply embedded in vS-Graphs and work closely with other modules to enrich the reconstructed maps and ensure optimal performance.

% Pipeline
In brief, RGB-D data is processed in real time, providing integrated visual and depth information to subsequent modules.
Visual features are extracted and tracked across frames in the \textit{Tracking} thread, where pose information is either initialized or refined, depending on the map reconstruction stage, thereby creating a 3D map of tracked features.
KeyFrame selection, a critical step following feature extraction, is performed within the \textit{Tracking} thread by analyzing visual cues.
These KeyFrames contain 3D map points and point clouds, forming the foundation for subsequent processes.
KeyFrames are then sent to the \textit{Local Mapping} thread for map integration and optimization, with inaccurately posed KeyFrames culled for enhanced accuracy.
Simultaneously, the \textit{Building Component Recognition} thread identifies and localizes walls and ground surfaces by processing the KeyFrame-level visual-spatial data.
They are detected by filtering the data through a \ac{CNN}, followed by RANSAC plane fitting for global pose estimation.
Concurrently, the \textit{Structural Element Recognition} thread runs at fixed intervals, extracting higher-level entities, including rooms and floors, from the building components of the current (active) map.
Eventually, the \textit{local bundle adjustment} module in the \textit{Local Mapping} thread jointly optimizes KeyFrame poses, map points, building components, and structural elements as coupled variables in the factor graph, ensuring that geometric and semantic entities are mutually refined for accurate map reconstruction.

At baseline, \textit{Atlas} is the core of the multi-map management, maintaining mapped instances and their associations and supporting operations such as map merging.
In vS-Graphs, we extend the module to store building components and structural elements, enriching map storage and retrieval beyond purely geometric information.
Furthermore, the \textit{Loop Closure Detection} checks for revisited locations and triggers \textit{Global Bundle Adjustment} to optimize and merge the map upon detection of a loop.
While it remains fundamentally geometric, the inclusion of environment-driven entities enables the system to verify and refine loop candidates.
During global optimization, redundant building components and structural elements detected across overlapping KeyFrames are merged and re-optimized within the graph, with their associations updated to ensure map consistency.
Although \textit{semantic loop closure} provides complementary validation, fully semantic-driven loop closure remains an area for future investigation.
\subsection{Building Component Recognition}
\label{sec_proposed_building}

% Intro
This thread processes the point cloud and visual data from KeyFrames to extract fundamental environment-driven elements beneficial for scene understanding.
The current version of vS-Graphs defines building components \(\mathbf{\Psi}\) as walls and ground surfaces.
Accordingly, each KeyFrame \(K=\{\mathbf{P}, L, \varepsilon\}\) passes through the \textit{building component recognition} module \(B\), generating identified structural element set \(\mathbf{\Pi}_\mathbf{\Psi}^K=B(\mathbf{K})\) for \(K\).
Here, \(\mathbf{P} = \{p_i ~|~ p \in \mathbb{R}^3, i \in \mathbb{N}\}\) is the point cloud, \(L\) is the matrix of RGB data, and \(\varepsilon\) represents additional metadata for mapping (such as camera pose).

% Semantic Segmentation
Building component extraction is achieved by processing \(L\) with a semantic segmentation function \(S\), which labels each pixel in the RGB data, retaining only relevant classes and discarding irrelevant ones.
In contrast with the conventional object detection methods, panoptic scene segmentation provides \textit{pixel-level class labels} and \textit{instance differentiation}, enabling sharper boundary delineation and more precise object recognition.
In principle, any reliable real-time panoptic segmentation framework can be integrated into vS-Graphs.
However, the current version integrates Panoptic-FCN (pFCN) \cite{pfcn} and YOSO \cite{yoso} by default, given their efficiency and strong balance between accuracy and real-time performance.
In this regard, \(\widehat{\mathbf{P}}_\mathbf{\Psi}^K=S(\mathbf{P}, L)\) refers to the segmented visual-spatial data of \(K\), where \(L_{(u,v)}\) is the semantic class label of a pixel at \((u,v) \in \mathbb{N}^2\).
Applying semantic class filtration to filter building components' semantic classes \(\mathbf{\Psi}\) (\textit{i.e.,} wall and ground surfaces) takes place as follows:
\begin{equation}
    \widehat{\mathbf{P}}_{\psi}^K = \{ p_j ~|~ p_j \in \mathbb{R}^3, L_{p_j} \in \mathbf{\Psi}\}
    \label{eq_semseg_pc}
\end{equation}
\begin{equation}
    \widehat{\mathbf{P}}_\mathbf{\Psi}^K = \{\widehat{\mathbf{P}}_{\psi}^K\} = \{\widehat{\mathbf{P}}_{wall}^K, \widehat{\mathbf{P}}_{ground}^K\}
    \label{eq_semseg_pcs}
\end{equation}
\noindent where \(\widehat{\mathbf{P}}_\mathbf{\Psi}^K \subset \mathbf{P}\) represents the set of semantically segmented point clouds, including the point clouds of \(\widehat{\mathbf{P}}_{wall}^K\) and \(\widehat{\mathbf{P}}_{ground}^K\).
It should be noted that the classification confidence \(\lambda_{(u,v)} \in \mathbb{R}\) for each pixel \({L_\Psi}_{(u,v)}\) is set for potential classification errors.

% PC Processing
The next stage is to optimize \(\widehat{\mathbf{P}}_\mathbf{\Psi}^K\), as it may contain noisy or low-resolution points that negatively impact subsequent steps.
In this regard, each segmented visual-spatial point cloud \(\widehat{\mathbf{P}}_{\psi}^K\) undergoes a two-stage preprocessing procedure.
The first step is applying down-sampling to fetch \(\widehat{\mathbf{P}}_{{\psi}_d}^K\), a refined point cloud with reduced redundancy and noise, where \(\widehat{\mathbf{P}}_{{\psi}_d}^K \subset \widehat{\mathbf{P}}_{\psi}^K\).
As depth sensors exhibit increased noise at extreme ranges, the semantic classification of structural elements is most reliable within an optimal depth range.
Thus, applying distance filtration based on the recommended sensor range results in retaining a subset of the point cloud as $\widehat{\mathbf{P}}_{{\psi}_{\zeta}}^K \subset \widehat{\mathbf{P}}_{{\psi}_d}^K$.
The final processed point cloud, referred to as \(\widehat{\mathbf{P}}_\mathbf{{\Psi}_{\zeta}}^K = \{\widehat{\mathbf{P}}_{{\psi}_{\zeta}}^K\}\), is then forwarded to successive stages for further analysis.

% RANSAC
Processing each \(\widehat{\mathbf{P}}_{{\psi}_{\zeta}}^K\) through \ac{RANSAC} \cite{ransac} plane fitting algorithm results in detecting semantically-validated building components with their geometric equations.
Thus, sets of random points \(\mathbf{p_r} \in \widehat{\mathbf{P}}_{{\psi}_{\zeta}}^K\) are iteratively selected to calculate the normal vectors \(\mathbf{n} \in \mathbb{R}^3\) representing validated planar components \(\mathbf{\pi}\) with the pre-defined distance \(d(\mathbf{{p_r}}, \pi) \leq \epsilon\), where \(\epsilon\) is the inlier threshold.
Accordingly, the final output representing all defined building components is as follows:
\begin{equation}
    \mathbf{\Pi}_\psi^K = \{{\pi_\psi}_i^K ~|~ {\pi_\psi}_i^K \in \mathbb{R}^3, i \in \mathbb{N}\}
    \label{eq_semseg_bc}
\end{equation}
\begin{equation}
    \mathbf{\Pi}_\mathbf{\Psi}^K = \{\mathbf{\Pi}_{\psi}^K\} = \{\mathbf{\Pi}_{wall}^K, \mathbf{\Pi}_{ground}^K\}
    \label{eq_semseg_bcs}
\end{equation}

% Post-processing
% Finally, planar surfaces that do not belong to wall or ground entities are omitted from \(\mathbf{\Pi}_\mathbf{\Psi}^K\), keeping only validated building components with estimated geometric equations.
The remaining reliable elements are checked against structural validation rules \(\mathbf{\Pi}_\mathbf{\Psi}^K = V(\mathbf{\Pi}_\mathbf{\Psi}^K)\), ensuring that they satisfy reasonable geometric constraints for the environment.
For instance, walls should be represented as vertical planes, while ground surfaces should be defined as horizontal planes.
The thread concludes the processing of KeyFrame \(K\) by storing its detected building components \(\mathbf{\Pi}_\mathbf{\Psi}^K\) within the current map in the \textit{Atlas}, ensuring they contribute to the ongoing map reconstruction.
\subsection{Structural Element Recognition}
\label{sec_proposed_structural}

% Intro
This thread repeatedly runs at constant time intervals (\textit{i.e.,} every two seconds) to detect potential higher-level semantic entities that characterize the environment’s layout, including rooms and floors.
Structural elements comprise multiple building components, with their topological associations taken into account.
The thread actively searches for layouts that form rooms, where a room is a spatially enclosed area bounded by at least two walls that surround a confined free-space cluster on the ground plane.
Additionally, a floor represents a higher-order structural element that encompasses a collection of rooms within a single building level.

% Associate BCs
In the first step, the map's existing building components \(\mathbf{\Pi}_\mathbf{\Psi} = \{\mathbf{\Pi}_\mathbf{\Psi}^{K_i} ~|~i\in \mathbb{N}\}\) are fetched from \textit{Atlas}.
Associating and merging redundant or conflicting building components is crucial to ensure consistency in the map reconstruction.
In this context, the primary factors to assess are the spatial proximity and alignment of the mapped building components, requiring evaluating both the Euclidean distance and the angular difference between their normal vectors.
Accordingly, the building component association condition is presented below:
\begin{equation}
    \| \mathbf{\Pi}_\mathbf{\Psi}^{K_i} - \mathbf{\Pi}_\mathbf{\Psi}^{K_j} \| \leq \rho \quad \wedge
    \quad \cos^{-1} \left( \frac{\mathbf{n}_i \cdot \mathbf{n}_j}{\|\mathbf{n}_i\| \|\mathbf{n}_j\|} \right) \leq \eta
    \label{eq_bc_associat}
\end{equation}
\noindent where $\mathbf{n}$ is the normal vector of a particular surface and $\rho$ and $\eta$ refer to spatial proximity and angular alignment thresholds, respectively.
The associated building components \(\mathbf{\Pi}_\mathbf{\Psi}\) are subsequently utilized to detect structural elements \(\mathbf{\Delta}_\mathbf{\Phi}\), defined as below:
\begin{equation}
    \mathbf{\Delta}_\phi^K = \{{\delta_\phi}_m^K ~|~ {\delta_\phi}_m^K \in \mathbb{R}^3, m \in \mathbb{N}\}
    \label{eq_str}
\end{equation}
\begin{equation}
    \mathbf{\Delta}_\mathbf{\Phi}^K = \{\mathbf{\Delta}_{\phi}^K\} = \{\mathbf{\Delta}_{room}^K, \mathbf{\Delta}_{floor}^K\}
    \label{eq_str_arr}
\end{equation}
% \begin{equation}
%     \mathbf{\Delta}_\mathbf{\Phi}^K = \{\mathbf{\Delta}_{\phi}^K\} = \{\mathbf{\Delta}_{corridor}^K, \mathbf{\Delta}_{room}^K, \mathbf{\Delta}_{floor}^K\}
%     \label{eq_str_arr}
% \end{equation}
\noindent where \(\mathbf{\Delta}_\mathbf{\Phi}^K \in \mathbf{\Delta}_\mathbf{\Phi}\) is the set of structural elements belong to semantic classes \(\mathbf{\Phi}\) found in KeyFrame \(K\).
The procedure for detecting structural elements is outlined below. \\

\textbf{Rooms.}
A room \(\delta_{r}^K = \{\mathbf{\Pi}_{wall}, {\pi_{ground}^{K_q}}, \mathbf{\nu}_r\}\) is a convex, spatially enclosed area bounded by at least two walls, each oriented toward a distinct free-space cluster $\mathbf{\Upsilon} = \{\upsilon_j ~|~ \upsilon_j \in \mathbb{R}^3, j \in \mathbb{N}\}$ in the global frame.
The free-space cluster is calculated from the depth camera points using the Voxblox tool \cite{voxblox}.
The associated set of walls is denoted as $\mathbf{\Pi}_{wall} = \{\pi_{wall}^{K_1}, \dots, \pi_{wall}^{K_n}\}$, where $n \geq 2$.
Each wall $\pi_{wall}^{K_i}$ is validated using \textit{proximity} and \textit{directionality} constraints, ensuring geometric consistency with its corresponding free-space.
\textit{Proximity} is calculated as:
\begin{equation}
    \forall\, \pi_{wall}^{K_i} \in \mathbf{\Pi}_{wall},\ 
    \exists\, \upsilon_j \in \mathbf{\Upsilon}:\ 
    \left|\, \mathbf{n}_{wall}^{K_i\top} \cdot \upsilon_j + d_{wall}^{K_i} \,\right| \le \omega
    \label{eq_wall_cluster_distance}
\end{equation}
where $\mathbf{n}_{wall}^{K_i} \in \mathbb{R}^3$ denotes the wall’s normal vector, $d_{wall}^{K_i}$ is the plane offset, and $\omega$ is a tunable threshold controlling the allowable distance between the wall and nearby free-space points.
\textit{Directionality} is computed as follows:
\begin{equation}
    \forall\pi_{wall}^{K_i} \in \mathbf{\Pi}_{wall}:\
    \mathbf{n}_{wall}^{K_i\top}
    \left(
    \mathbf{\nu}_{\Upsilon} - \mathbf{\nu}_{wall}^{K_i}
    \right)
    < 0
    \label{eq_wall_cluster_directionality}
\end{equation}
where $\mathbf{\nu}_{wall}^{K_i} \in \mathbb{R}^3$ is the wall centroid and $\mathbf{\nu}_{\Upsilon} \in \mathbb{R}^3$ is the centroid of the associated free-space cluster $\Upsilon$.
These formulations enable the detection of convex $n$-wall rooms with arbitrary wall orientations, beyond Manhattan layouts, by enforcing that adjacent walls face the enclosed cluster.
The ground plane $\pi_{ground}^{K_q}$ is associated with a room if it is approximately orthogonal to all wall normals and spatially enclosed by them.
Accordingly:
\begin{equation}
    \left| \mathbf{n}_{ground}^\top \cdot \mathbf{n}_{wall}^{K_i} \right| \leq \sin(\vartheta) 
    \quad \wedge \quad 
    d_{\pi_{wall}^{K_i}} \leq \nu_g \leq d_{\pi_{wall}^{K_j}}
    \label{eq_ground_orthogonality_short}
\end{equation}
\noindent where $\vartheta$ defines the angular tolerance and $\nu_g$ refers to the ground centroid lying between the nearest and farthest walls.

% Cost
The optimization incorporates a geometry-aware constraint mechanism that updates each room’s centroid ($\mathbf{\nu}{r_i}$) based on the centroids of its associated walls, while simultaneously identifying \textit{parallel} and \textit{perpendicular} wall pairs within the same room.
Such patterns are naturally common in man-made environments, where walls are commonly constructed to align along dominant structural axes.
These relationships introduce geometric constraints that minimize angular deviations, encouraging walls to align closer to $0^\circ$ or $90^\circ$, respectively.
In the absence of such relations, only the centroid consistency term remains active, maintaining spatial coherence without enforcing unnecessary structure.
This flexible design enables the system to accommodate rooms of arbitrary or asymmetric geometry without relying on prior layout assumptions, unlike rectangular-shaped formulations commonly used in previous works \cite{sgraphs, hydra, hovsg}.
In this context, if a room contains parallel walls facing each other toward $\mathbf{\nu}_{\Upsilon}$, a parallelism constraint penalizes deviation through a cost given by the angle between their normals:
\begin{equation}
    {c_{\delta_{r}^K}^{\parallel}}_{(i,j)} = 1 - \left| \mathbf{n}_i^\top \cdot \mathbf{n}_j \right|,
    \quad \text{where } |\mathbf{n}_i| = |\mathbf{n}_j| = 1.
\end{equation}
\noindent where $\mathbf{n}_i$ and $\mathbf{n}_j$ are the normalized values of normal vectors of the two wall planes.
The cost reaches zero for perfectly parallel walls ($\mathbf{n}_i \parallel \mathbf{n}_j$) and increases proportionally with angular deviation, providing a numerically stable measure of parallelism.
Similarly, perpendicularity is calculated as:
\begin{equation}
    {c_{\delta_{r}^K}^{\perp}}_{(i,j)} = \left| \mathbf{n}_i^\top \cdot \mathbf{n}_j \right|,
    \quad \text{where } |\mathbf{n}_i| = |\mathbf{n}_j| = 1.
\end{equation}
Finally, a centroid consistency term $c_{\mathbf{\nu}_r}$ keeps the room centroid $\mathbf{\nu}_r$ aligned with the mean of its wall centroids.
The overall cost function for room-level constraints is:
\vspace{-0.5em}
\begin{equation}
    c_{\mathbf{\nu}_r} = \| \hat{\mathbf{\nu}_r} - \frac{1}{n}\sum_{j=1}^{n}\mathbf{\nu}_{wall}^{K_j} \|^2
\end{equation}
\vspace{-1em}
\begin{equation}
    c_{\delta_{r}^K} =
    \frac{1}{N_{\parallel}} \sum_{i,j=0}^{N_{\parallel}} {c_{\delta_{r}^K}^{\parallel}}_{(i,j)}
    + \frac{1}{N_{\perp}} \sum_{i,j=0}^{N_{\perp}} {c_{\delta_{r}^K}^{\perp}}_{(i,j)}
    + c_{\mathbf{\nu}_{r_i}}
\end{equation}
\noindent were $N_{\parallel}$ and $N_{\perp}$ denote the number of parallel and perpendicular walls.
Hence, when no parallel or perpendicular relations are detected, only the Euclidean-norm–based centroid update is activated, preserving localization accuracy while enhancing the structural coherence of the 3D scene graph.
The factor weights were determined empirically by evaluating different configurations on a subset of sequences and selecting values that balance structural regularization with reprojection accuracy. These weights are fixed across all datasets and sequences without per-dataset tuning. All weight values are provided in the open-source implementation to facilitate reproducibility. \\ \\
\textbf{Floors.}
A floor $\delta_{f}^K = \{\mathbf{\Delta}_r, \mathbf{\nu}_f, \pi_{floor}^{K_i}\}$ is defined by a set of rooms $\mathbf{\Delta}_r$, all sharing a common horizontal reference plane $\pi_{floor}^{K_i}$ and the floor centroid $\mathbf{\nu}_f$.
The floor plane is associated with the constituent rooms if it satisfies the co-planarity condition within a vertical tolerance.
In practice, this validation step is ignored in the current vS-Graphs implementation due to its single-floor design constraint.
Additionally, its spatial extent is bounded by the union of all constituent room boundaries.
The floor centroid $\mathbf{\nu}_f$ is computed as the weighted arithmetic mean of all constituent room centroids:
\vspace{-0.5em}
\begin{equation}
    \mathbf{\nu}_f = \frac{\sum_{j=1}^{N_r} \mathbf{\nu}_{r_j}}{N_r}
    \label{eq_floor_centroid}
\end{equation}
\noindent where $N_r = |\mathbf{\Delta}_{room}|$.
The cost function to optimize the floor's vertex node is calculated as follows:
\setlength{\abovedisplayskip}{4pt}
\begin{equation}
    c_{\delta_f^K} = \sum_{t=1}^{T} \big\| \hat{\mathbf{\nu}}_f - h(\mathbf{\Delta}^K_r) \big\|^2_{\mathbf{\Lambda}_{\tilde{\delta}_{f,t}}}
    \label{eq_floor_node}
\end{equation}
\noindent where $h(\dots)$ is a hierarchical mapping function that associates the floor's structural components with its centroid.
\begin{figure}[!ht]
    \centering
    \includegraphics[width=0.95\textwidth]{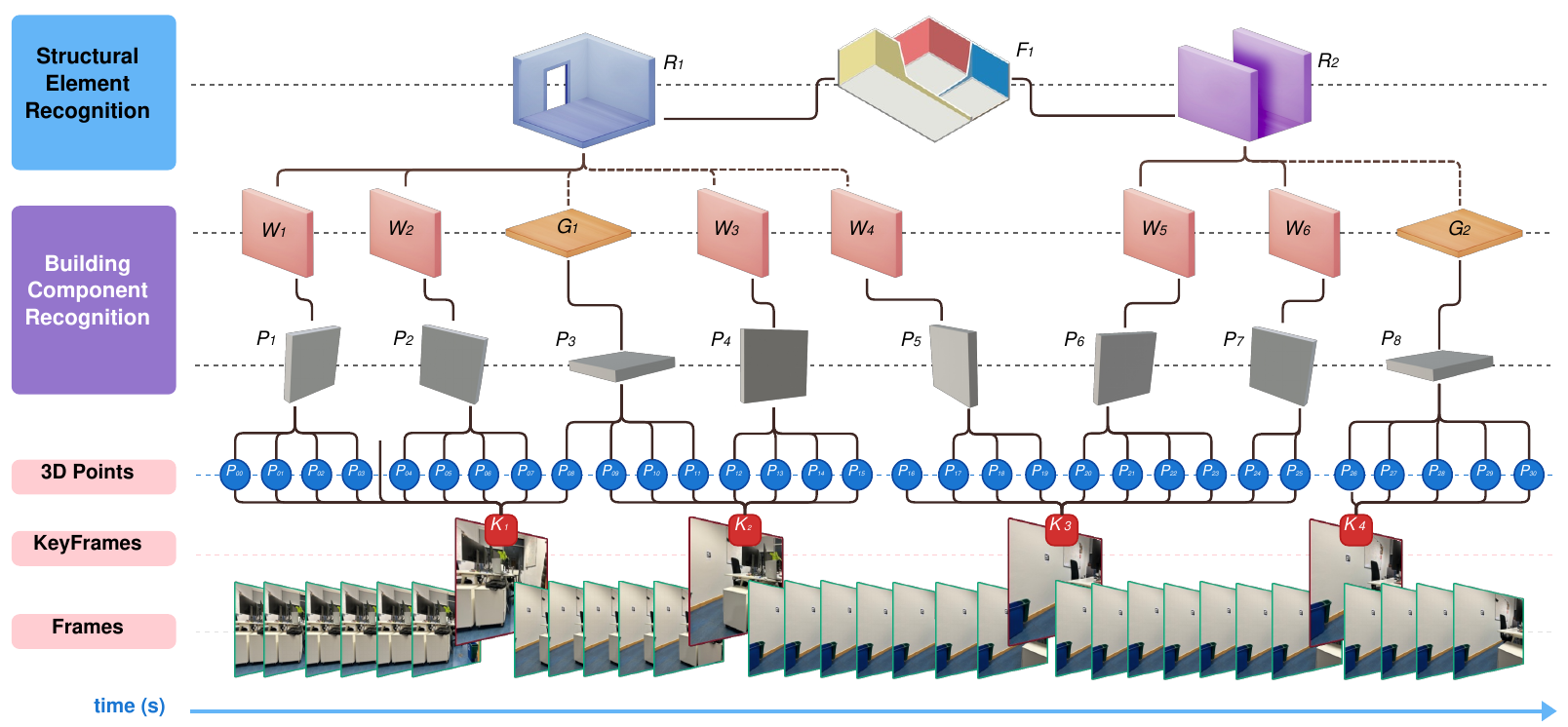}
    \caption{Scene graph structure generated using vS-Graphs, creating a hierarchical representation of the environment.}
    \label{fig_graph}
\end{figure}

\subsection{Scene Graph Structure}
\label{sec_proposed_graph}

% Intro
Fig.~\ref{fig_graph} illustrates the 3D scene graph structure generated using vS-Graphs, along with its corresponding modules shown in Fig.~\ref{fig_flowchart}.
In contrast to the traditional \ac{SLAM} reconstructed maps, the geometric replicas of vS-Graphs are augmented with hierarchical, rich semantic data, enabling meaningful interaction with the environment.
The generated scene graph fills the contextual understanding gap and provides better scene interpretation, enabling complementary missions such as scalable map-building and improved decision-making.
\section{Experimental Results}
\label{sec_evaluation}

\subsection{Evaluation Criteria}
\label{sec_eval_setup}

\begin{table}[t]
    \centering
    \caption{
        Characteristics of the \textit{AutoSense} in-house dataset. 
        Each sequence includes one ground and one floor surface by design. 
        \textit{BC} denotes building components (\textit{ground/walls}), 
        and \textit{SE} denotes structural elements (\textit{rooms/floor}).
    }
    \begin{tabular}{l|c|c|l}
        \toprule
            \multirow{2}{*}{\textbf{Seq.}} & \textbf{\#BC} & \textbf{\#SE} & \multirow{2}{*}{\textbf{Description}} \\
        % \cmidrule{2-3}
             & {\fontsize{6}{6}\selectfont \textit{(Ground+Walls)}} 
             & {\fontsize{6}{6}\selectfont \textit{(Room+Floor)}} & \\
        \midrule
            SR01 & 4 (1+3) & 2 (1+1) & partial view of a room \\
            SR02 & 4 (1+3) & 2 (1+1) & room corner with intersecting walls \\
            SR03 & 6 (1+5) & 2 (1+1) & a single rectangular room \\
        \midrule
            MR01 & 14 (1+13) & 4 (3+1) & across rooms connected to a corridor \\
            MR02 & 13 (1+12) & 4 (3+1) & adjoining rooms linked by a corridor \\
            MR03 & 22 (1+21) & 6 (5+1) & a suite of interconnected rooms \\
        \bottomrule
    \end{tabular}
    \label{tbl_dataset}
\end{table}

\begin{figure}[!ht]
    \centering
    \includegraphics[width=0.6\columnwidth]{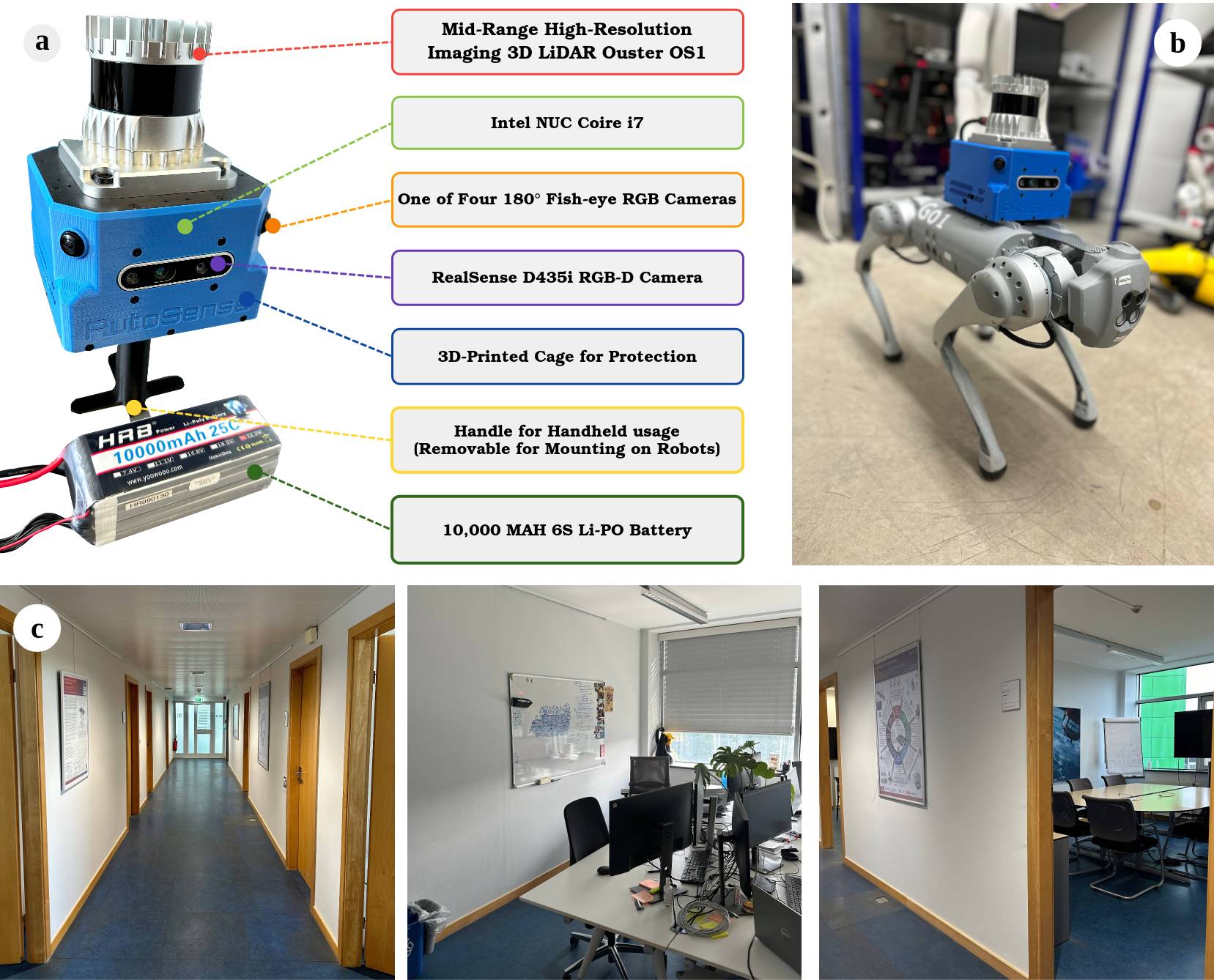}
    \caption{In-house dataset collection using the \textit{SMapper} device: \textbf{a)} the setup overview, \textbf{b)} the device mounted on a legged robot, and \textbf{c)} some instances of the collected data.}
    \label{fig_dataset}
\end{figure}

% Setup
\noindent \textbf{Setup.}
Evaluations were conducted on a system with an Intel\textregistered{} Core\texttrademark{} i9-11950H processor (2.60 GHz), a 4GB NVIDIA T600 Mobile GPU, and 32GB of RAM.

% Data
\noindent \textbf{Datasets.}
We evaluated vS-Graphs on standard benchmarks and in-house datasets to assess its performance across diverse environments and scene complexities.
The standard datasets include ICL \cite{dataset_icl} photorealistic synthetic scenes and real-world data from OpenLORIS \cite{dataset_openloris}, ScanNet \cite{dataset_scannet}, and TUM-RGBD \cite{dataset_tum}.
The in-house dataset, SMapper, was collected using a custom-built device introduced in \cite{smapper} that records RGB-D video and \ac{LiDAR} point clouds.
It features diverse real-world indoor environments with varied architectural layouts and multi-room scenarios.
The ground-truth data in the dataset were derived from reliable LiDAR poses and point clouds generated by \sgraphs.
Table \ref{tbl_dataset} outlines the dataset characteristics, encompassing approximately $44~\mathrm{minutes}$ of operation and $620~\mathrm{meters}$ of trajectories across diverse scenarios.
Additionally, Fig.~\ref{fig_dataset} showcases the device and sample sequences.
Due to space constraints, the complete evaluation results and figures are available at \href{https://snt-arg.github.io/vsgraphs-results/}{https://snt-arg.github.io/vsgraphs-results/}.
\subsection{Evaluations and Discussions}
\label{sec_eval}

\begin{table*}[!ht]
    \scriptsize
    \fontsize{7.5}{7.0}\selectfont % {fontSize}{spaceSize}
    \setlength{\tabcolsep}{2pt}
    \centering
    \caption{\acf{ATE} in centimeters ($\mathrm{cm}$) for various \ac{VSLAM} algorithms across eight iterations.
    The best and second-best results are \textbf{boldfaced} and \underline{underlined}, respectively.
    vS-Graphs is evaluated with multiple semantic segmentation backbones (YOSO and pFCN) and with different levels of entity integration, including only building component cues (\textit{vS-Graphs BC}) and the full pipeline with structural element recognition (\textit{vS-Graphs}).
    The minimum completion rate of each sequence is set to 50\%, and dashes indicate unavailable data due to tracking failures.
    Additionally, the ``\textit{\%Gain}'' rows quantify the percentage improvement achieved by vS-Graphs compared to its baseline (ORB-SLAM 3.0).}
    \resizebox{\textwidth}{!}{
    \begin{tabular}{l | l | c | c | cccc | cc | cc | cc | cc}
    \toprule
        \multirow{6}{*}{\textbf{Dataset}} & \multirow{6}{*}{\textbf{Sequence}} & \multirow{6}{*}{\textbf{Time (s)}} & \multirow{6}{*}{\textbf{Length (m)}} & \multicolumn{8}{c}{\textbf{Methodology}} \\
    \cmidrule(lr){4-16}
        & & & & \multirow{3}{*}{\rotatebox{45}{\tiny\textbf{ORB-SLAM3}}} & \multirow{3}{*}{\rotatebox{45}{\tiny\textbf{BAD SLAM}}} & \multirow{3}{*}{\rotatebox{45}{\tiny\textbf{ElasticFusion}}} & \multirow{3}{*}{\rotatebox{45}{\tiny\textbf{DROID-SLAM}}} & \multicolumn{4}{c|}{\textbf{vS-Graphs}} & \multicolumn{4}{c}{\textbf{\textit{\%Gain}}} \\
    \cmidrule(lr){9-16}
         & & & & & & & & \multicolumn{2}{c|}{\textit{BC Only}} & \multicolumn{2}{c|}{\textit{BC+SE}} & \multicolumn{2}{c|}{\textit{BC Only}} & \multicolumn{2}{c}{\textit{BC+SE}} \\
    \cmidrule(lr){9-16}
        & & & & & & & & YOSO & pFCN & YOSO & pFCN & YOSO & pFCN & YOSO & pFCN \\
    \midrule
        \multirow{6}{*}{\rotatebox{45}{\textbf{ICL}}} & deer-ground & 79.9 & 11.7 & \textbf{0.74} & 147.60 & 14.50 & 0.90 & 0.81 & 0.91 & \underline{0.78} & 0.89 & -8.89 & -23.40 & -5.25 & -20.09 \\
        & deer-walkh & 65.3 & 34.6 & 6.96 & -- & 82.50 & 44.64 & \underline{6.08} & 6.23 & \textbf{5.96} & 6.14 & 12.64 & 10.54 & 14.38 & 11.80 \\
        & deer-walk & 64.0 & 45.1 & 9.91 & 147.40 & 62.00 & 2.87 & 8.82 & 12.19 & \textbf{2.44} & \underline{4.02} & 10.97 & -22.98 & 75.38 & 59.43 \\
        & deer-running & 28.4 & 16.0 & 6.92 & -- & 78.70 & -- & 5.34 & 5.47 & \textbf{2.83} & \underline{3.34} & 22.82 & 20.98 & 59.12 & 51.75 \\
        & deer-mav-fast & 102.5 & 74.9 & 2.67 & 4.60 & -- & 10.39 & 2.65 & \underline{2.07} & 2.23 & \textbf{1.26} & 0.96 & 22.57 & 16.61 & 52.88 \\
        \cmidrule(lr){2-16}
        & \textbf{Total} & 340 & 182.3 & 5.44 & 99.87 & 59.43 & 14.70 & 4.74 & 5.37 & \textbf{2.85} & \underline{3.13} & 7.70 & 1.54 & 32.05 & 31.15 \\
    \midrule
    \midrule
        \multirow{9}{*}{\rotatebox{45}{\textbf{OpenLORIS}}} & office1-1 & 27.0 & 5.3 & \underline{6.37} & 12.30 & 6.88 & 8.65 & 6.41 & \textbf{6.33} & 6.42 & \underline{6.37} & -0.66 & 0.50 & -0.85 & -0.06 \\
        & office1-2 & 30.2 & 5.9 & 10.05 & 14.00 & 11.26 & 11.12 & 9.91 & 9.93 & \textbf{9.28} & \underline{9.68} & 1.39 & 1.24 & 7.67 & 3.69 \\
        & office1-3 & 12.1 & 0.4 & 14.73 & 19.50 & -- & 16.75 & \underline{14.57} & 16.05 & \textbf{14.44} & 15.01 & 1.08 & -9.00 & 1.94 & -1.93 \\
        & office1-4 & 29.4 & 6.1 & 12.51 & 34.30 & 18.03 & 16.96 & 12.68 & 12.55 & \underline{11.54} & \textbf{11.36} & -1.37 & -0.33 & 7.77 & 9.21 \\
        & office1-5 & 52.6 & 10.5 & 11.57 & 30.20 & 22.15 & 24.94 & 11.11 & 11.29 & \underline{10.71} & \textbf{10.59} & 4.03 & 2.48 & 7.45 & 8.49 \\
        & office1-6 & 36.5 & 5.5 & 5.86 & 10.70 & -- & 8.14 & 5.90 & 5.89 & \underline{5.70} & \textbf{5.67} & -0.68 & -0.45 & 2.76 & 3.27 \\
        & office1-7 & 38.6 & 6.1 & 17.17 & 20.30 & -- & 11.28 & 17.55 & 17.50 & \textbf{13.00} & \underline{13.25} & -2.17 & -1.92 & 24.30 & 22.84 \\
        \cmidrule(lr){2-16}
        & \textbf{Total} & 226.4 & 39.8 & 11.18 & 20.19 & 14.58 & 15.61 & 11.16 & 11.36 & \textbf{10.16} & \underline{10.28} & 0.23 & -1.07 & 7.29 & 6.50 \\
    \midrule
    \midrule
        \multirow{6}{*}{\rotatebox{45}{\textbf{ScanNet}}} & scn0041\_01 & 75.5 & 25.5 & 14.26 & 22.20 & 21.96 & 14.87 & 14.29 & 14.06 & \textbf{13.56} & \underline{13.78} & -0.27 & 1.37 & 4.88 & 3.34 \\
        & scn0200\_00 & 37.3 & 5.9 & 4.88 & -- & -- & -- & 4.67 & 4.65 & \underline{4.55} & \textbf{4.42} & 4.36 & 4.79 & 6.85 & 9.52 \\
        & scn0614\_01 & 36.2 & 13.3 & 13.55 & 13.90 & 20.23 & 12.37 & 12.50 & 13.38 & \textbf{12.36} & \underline{12.45} & 7.75 & 1.26 & 8.76 & 8.10 \\
        & scn0626\_00 & 21.1 & 6.1 & 9.60 & 27.30 & 18.91 & 8.49 & 9.50 & 9.22 & \underline{8.73} & \textbf{8.71} & 1.05 & 4.02 & 9.07 & 9.28 \\
        \cmidrule(lr){2-16}
        & \textbf{Total} & 170.1 & 50.8 & 10.57 & 21.13 & 20.36 & 11.91 & 10.24 & 10.33 & \textbf{9.80} & \underline{9.84} & 3.22 & 2.86 & 7.39 & 7.56 \\
    \midrule
    \midrule
        \multirow{7}{*}{\rotatebox{45}{\textbf{TUM-RGBD}}} & frb1-desk & 23.8 & 9.3 & 2.07 & 2.20 & 2.55 & \textbf{1.80} & 1.98 & \underline{1.96} & 2.04 & 2.03 & 4.25 & 5.18 & 1.44 & 1.93 \\
        & frb1-desk2 & 25.1 & 9.9 & 3.23 & 2.90 & 7.87 & \textbf{2.46} & 3.19 & 3.19 & 2.89 & \underline{2.85} & 1.17 & 1.21 & 10.50 & 11.74 \\
        & frb1-room & 49.1 & 16.0 & 13.25 & 20.90 & 16.75 & \textbf{3.76} & 13.02 & 13.29 & \underline{5.13} & 6.77 & 1.75 & -0.26 & 61.29 & 48.91 \\
        & frb2-desk-prs & 142.1 & 18.8 & 1.90 & 9.60 & 4.51 & 6.22 & \underline{1.79} & \textbf{1.78} & \underline{1.79} & 1.81 & 5.69 & 6.07 & 5.80 & 4.75 \\
        & frb3-strct & 31.9 & 6.0 & 1.62 & 2.10 & 1.75 & 1.74 & 1.54 & \underline{1.53} & \textbf{1.52} & 1.54 & 5.04 & 5.64 & 6.32 & 5.08 \\
        \cmidrule(lr){2-16}
        & \textbf{Total} & 272 & 50.7 & 4.41 & 7.54 & 6.68 & 3.20 & 4.31 & 4.35 & \textbf{2.67} & \underline{3.00} & 3.58 & 3.57 & 17.07 & 14.48 \\
    \midrule
    \midrule
        \multirow{8}{*}{\rotatebox{45}{\textbf{SMapper}}} & SR01 & 61.6 & 8.5 & 7.98 & 13.00 & \textbf{6.62} & 8.03 & 8.10 & 8.16 & \underline{7.34} & 7.63 & -1.58 & -2.28 & 7.99 & 4.36 \\
        & SR02 & 80.4 & 11.9 & 8.75 & 13.50 & -- & 9.38 & \textbf{8.21} & 8.29 & \underline{8.23} & 8.26 & 6.15 & 5.32 & 5.95 & 5.60 \\
        & SR03 & 170.4 & 36.2 & 10.90 & 55.70 & -- & 12.70 & \underline{10.70} & 10.71 & \textbf{10.51} & 11.04 & 1.85 & 1.71 & 3.56 & -1.30 \\
        & MR01 & 305.7 & 71.8 & 15.26 & 41.30 & -- & 14.88 & 15.44 & 14.98 & \textbf{12.23} & \underline{12.75} & -1.13 & 1.85 & 19.87 & 16.47 \\
        & MR02 & 210.2 & 51.7 & 23.97 & -- & 53.48 & 2406 & 21.68 & 23.78 & \textbf{20.13} & \underline{21.19} & 9.58 & 0.83 & 16.03 & 11.61 \\
        & MR03 & 783.3 & 135.5 & 22.19 & 81.21 & -- & 19.03 & 22.00 & 19.23 & \textbf{17.70} & \underline{17.84} & 0.87 & 13.35 & 20.24 & 19.61 \\
        \cmidrule(lr){2-16}
        & \textbf{Total} & 1611.6 & 295.2 & 14.84 & 40.94 & 30.05 & 16.01 & 14.35 & 14.19 & \textbf{12.69} & \underline{13.12} & 2.63 & 3.46 & 12.27 & 9.39 \\
    \midrule
    \midrule
        \multicolumn{2}{l|}{\textbf{Overall Total}} & 43m 40s & 618.8m & 9.29 & 37.93 & 26.22 & 12.29 & 8.96 & 9.12 & \textbf{7.63} & \underline{7.87} & 3.47 & 2.07 & 15.22 & 13.82 \\
    \bottomrule
    \end{tabular}
    }
    \label{tbl_eval}
\end{table*}

% Hint: A01=SR01, A02=SR02, A05=SR03, A06=MR01, A07=MR02, A08=MR03
% Hint [old]: A01=room-walls, A02=room-corner, A05=single-room, A06=room-across, A07=room-corrdr, A08=office-suite

\subsubsection{\textbf{Trajectory Estimation Performance}}
\label{sec_eval_slam}

To assess trajectory estimation accuracy, vS-Graphs was benchmarked against established and robust VSLAM systems, including ORB-SLAM 3.0 (baseline) \cite{orbslam3}, DROID-SLAM \cite{droidslam}, ElasticFusion \cite{elasticfusion}, and BAD SLAM \cite{badslam}.
Marker-based methods (\textit{e.g.,} \cite{semuco}) were excluded due to their reliance on artificial landmarks and external pose constraints, which limit their applicability in marker-free environments.
Similarly, neural field-based approaches (such as \cite{niceslam}) were omitted because they rely on learned scene priors rather than explicit geometric–semantic representations.
Additionally, vS-Graphs was evaluated using different segmentation backbones (pFCN and YOSO) to assess the effects of segmentation quality and efficiency on trajectory robustness.

% Trajectory
Table~\ref{tbl_eval} presents the evaluation results, where each system is evaluated over eight runs per sequence, and performance is measured using \acf{ATE} reported in centimeters.
Accordingly, vS-Graphs variants consistently achieve state-of-the-art performance, ranking first or second in nearly all evaluations across segmentation backbones and module configurations.
This improvement stems from the additional geometric and semantic constraints imposed by the accurate localization of building components and structural elements.
When combined with loop closures, these constraints yield even larger gains, \textit{e.g.,} $75.38\%$ for ``\textit{deer-w}'' with the YOSO-based configuration.
However, rapid motion and noisy depth data can negatively affect performance (\textit{e.g.,} ``\textit{SR01}'', BC-only).
DROID-SLAM performs better in rapid-motion scenarios (TUM-RGBD instances), but at the cost of substantially higher GPU memory usage and limited real-time capability.
Another notable observation is that incorporating room entities further improves trajectory accuracy, particularly when room–wall constraints are enforced.
This effect is most evident in looped (``\textit{deer-w}'') and multi-room environments (``\textit{MR01}''), where maintaining geometric consistency is crucial, reducing ATE by $13.82\%$ and $15.22\%$ w.r.t. the baseline.
It can be seen that the choice of panoptic backbone (YOSO or pFCN) has only a marginal effect on the trajectory accuracy of vS-Graphs, with higher segmentation quality naturally yielding more precise structural mapping.

% \input{figures/fig_discussion}

% Limitations
It should be noted that a primary challenge shared by all tested VSLAMs arises in low-texture scenes (\textit{e.g.,} corridors with uniformly painted walls), negatively impacting feature-matching and tracking procedures.
In vS-Graphs, this can affect pose estimation and localization of building components and propagate the error through the structural element recognition stage.
Additionally, treating building components as planar surfaces might limit the framework's applicability in environments with irregular geometries (\textit{e.g.,} curved walls). % in Fig.~\ref{fig_eval_rec_curve}
Furthermore, overly or loosely permissive association thresholds can degrade building component identification performance, making careful threshold selection crucial.

% Mapping
\subsubsection{\textbf{Mapping Performance}}
\begin{figure}[!t]
     \centering
     \includegraphics[width=0.9\columnwidth]{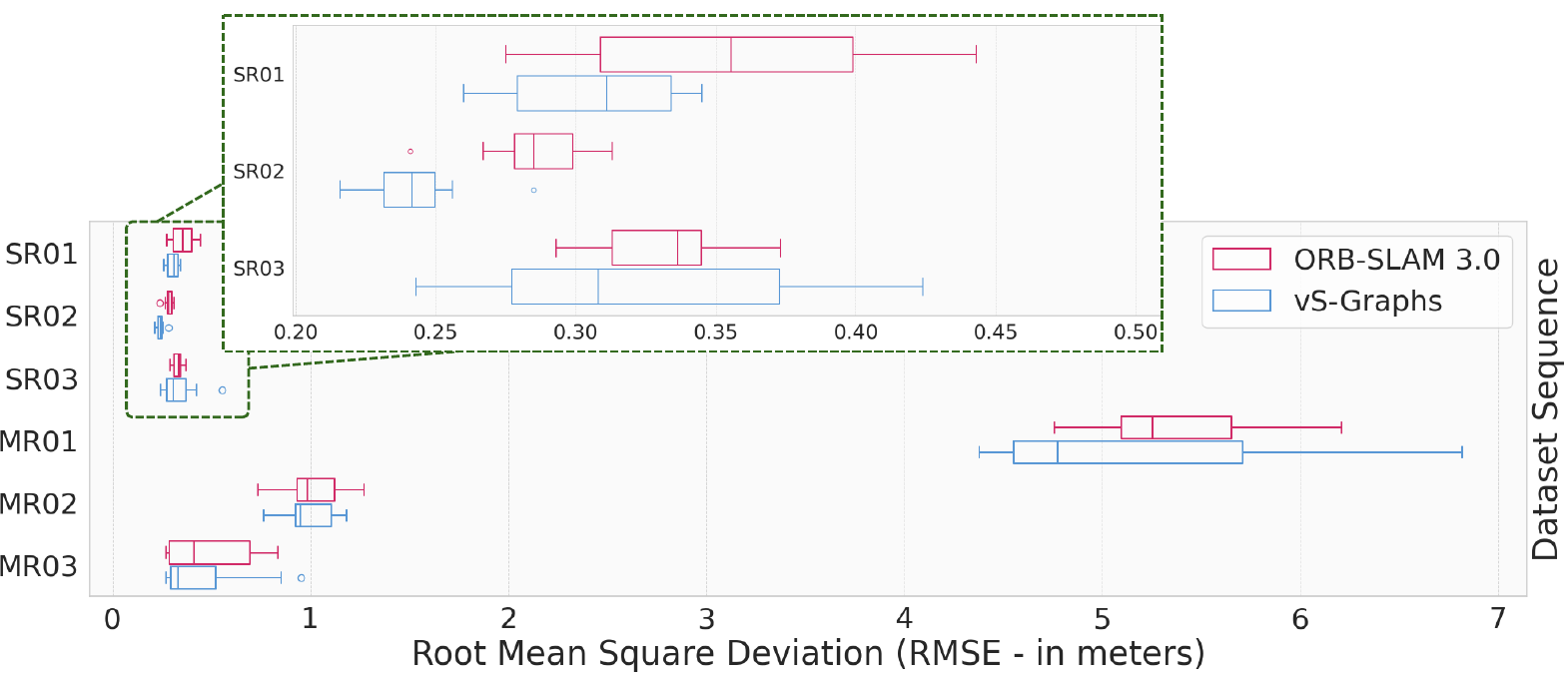}
     \caption{Mapping performance across eight iterations, showing SMapper sequences with less than one meter \ac{RMSE}.}
     \label{fig_eval_rmse}
\end{figure}

Additionally, analyzing the accuracy of the reconstructed maps against SMapper's ground truth data revealed that \vgraphs~performs more robustly compared to \orb~in terms of \ac{RMSE}.
As shown in Fig.~\ref{fig_eval_rmse}, the median \ac{RMSE} is consistently lower in \vgraphs, indicating a higher level of overall mapping precision.
\vgraphs~achieves superior mapping accuracy despite generating maps with $\sim10.15\%$ fewer points on average than the baseline, owing to its environment-driven constraints that enable a more coherent reconstruction.

\subsubsection{\textbf{Scene Understanding Performance}}
\label{sec_eval_semantic}

% Intro
This experiment evaluates vS-Graphs in terms of semantic scene understanding, with a particular focus on detecting entities essential for interpreting the environment’s structural layout. 
For benchmarking, we selected the \emph{multi-room sequences} from the \textit{AutoSense} dataset, which include ground truth annotations derived from LiDAR scans. 
Table~\ref{tbl_eval_rec} reports a quantitative comparison of vS-Graphs against two state-of-the-art approaches: Hydra \cite{hydra} and \textit{S-Graphs} \cite{sgraphsp}.
For presentation clarity, \textit{precision} and \textit{recall} values for structural elements were omitted from the table to avoid visual saturation.
The ground truth counts of rooms and walls were obtained by manual inspection of the dataset, where each entity was visually verified and counted.
While \textit{S-Graphs} benefits from the geometric accuracy of LiDAR point clouds, Hydra was configured to use visual point clouds, ensuring a fair comparison against our purely vision-based approach.

\begin{table}[!b]
    \centering
    \small
    \caption{Scene understanding accuracy of vS-Graphs on multi-room sequences of SMapper. \textit{BC} and \textit{SE} denote “building components” and “structural elements,” respectively, while \textit{GT} indicates the ground truth count of manually verified items.}
    \begin{tabular}{c|c|cc|c|c}
        \toprule
            \multicolumn{2}{c|}{} & \multicolumn{2}{c|}{\textbf{Detected / GT}} & \multicolumn{1}{c|}{\textbf{Precision}} &\multicolumn{1}{c}{\textbf{Recall}} \\
        \midrule
            \textit{Method} & \textit{Sequence} & \textit{BC} & \textit{SE} & \textit{BC} & \textit{BC} \\
        \midrule
            \multirow{3}{*}{{\shortstack{\textbf{\sgraphs} \\ \cite{sgraphsp}}}} & MR01 & 11 / 14 & 4 / 4 & \cellcolor{greenl}{0.92} & \cellcolor{greenl}{0.92} \\
            & MR02 & 12 / 13 & 4 / 4 & \cellcolor{greend}{1.00} & \cellcolor{greenl}{0.92} \\
            & MR03 & 20 / 22 & 6 / 6 & \cellcolor{greenl}{0.90} & \cellcolor{greend}{0.95} \\
        \midrule
            \multirow{3}{*}{{\shortstack{\textbf{Hydra} \\ \cite{hydra}}}} & MR01 & N/A & 4 / 4 & N/A & N/A \\
            & MR02 & N/A & 5 / 4 & N/A & N/A \\
            & MR03 & N/A & 6 / 6 & N/A & N/A \\
        \midrule
            \multirow{3}{*}{{\shortstack{\textbf{\vgraphs} \\ (ours)}}} & MR01 & 13 / 14 & 4 / 4 & \cellcolor{greenl}{0.86} & \cellcolor{greend}{1.00} \\
            & MR02 & 13 / 13 & 4 / 4 & \cellcolor{greenl}{0.92} & \cellcolor{greenl}{0.92} \\
            & MR03 & 23 / 22 & 6 / 6 & \cellcolor{greend}{0.96} & \cellcolor{greend}{1.00} \\
        \bottomrule
    \end{tabular}
    \label{tbl_eval_rec}
\end{table}

\begin{figure}[!t]
     \centering
     \begin{subfigure}[t]{0.35\columnwidth}
         \centering
         \includegraphics[width=\columnwidth]{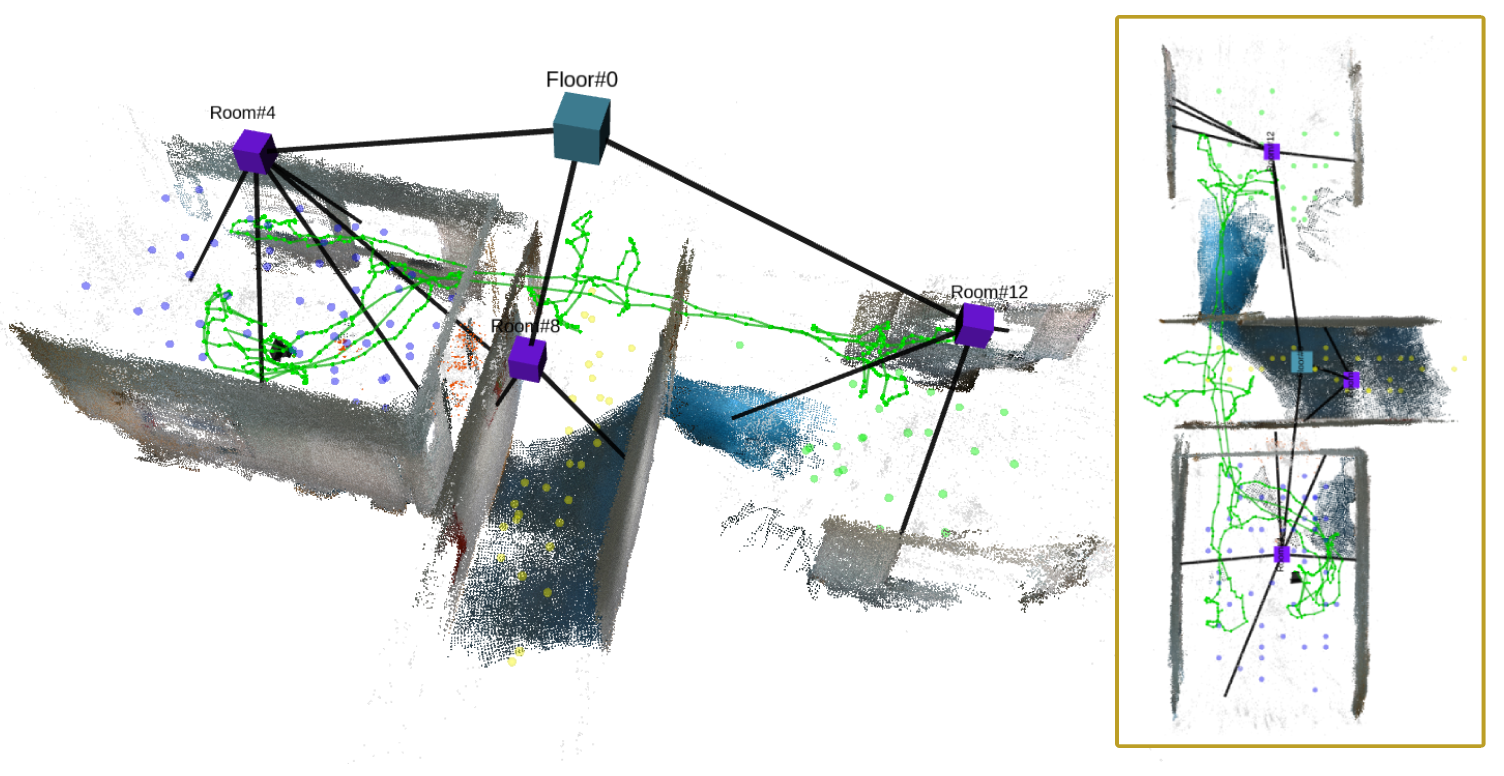}
         \caption{vS-Graphs on \textit{MR01}}
         \label{fig_eval_slam_scan_vgraph}
     \end{subfigure}
     \begin{subfigure}[t]{0.35\columnwidth}
         \centering
         \includegraphics[width=\columnwidth]{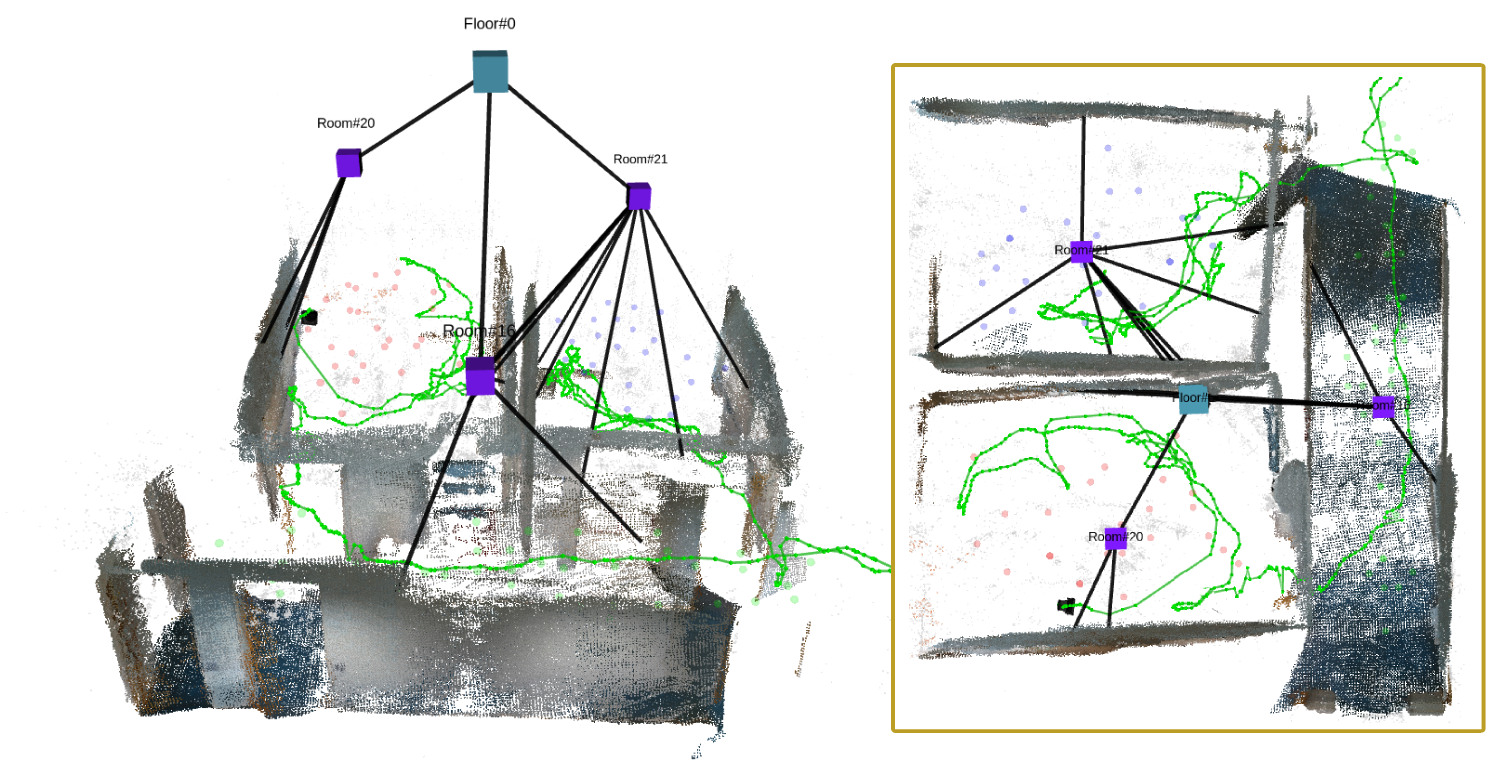}
         \caption{\vgraphs~on \textit{MR02}}
         \label{fig_eval_slam_lux_vgraph}
     \end{subfigure}
     \begin{subfigure}[t]{0.35\columnwidth}
         \centering
         \includegraphics[width=0.98\columnwidth]{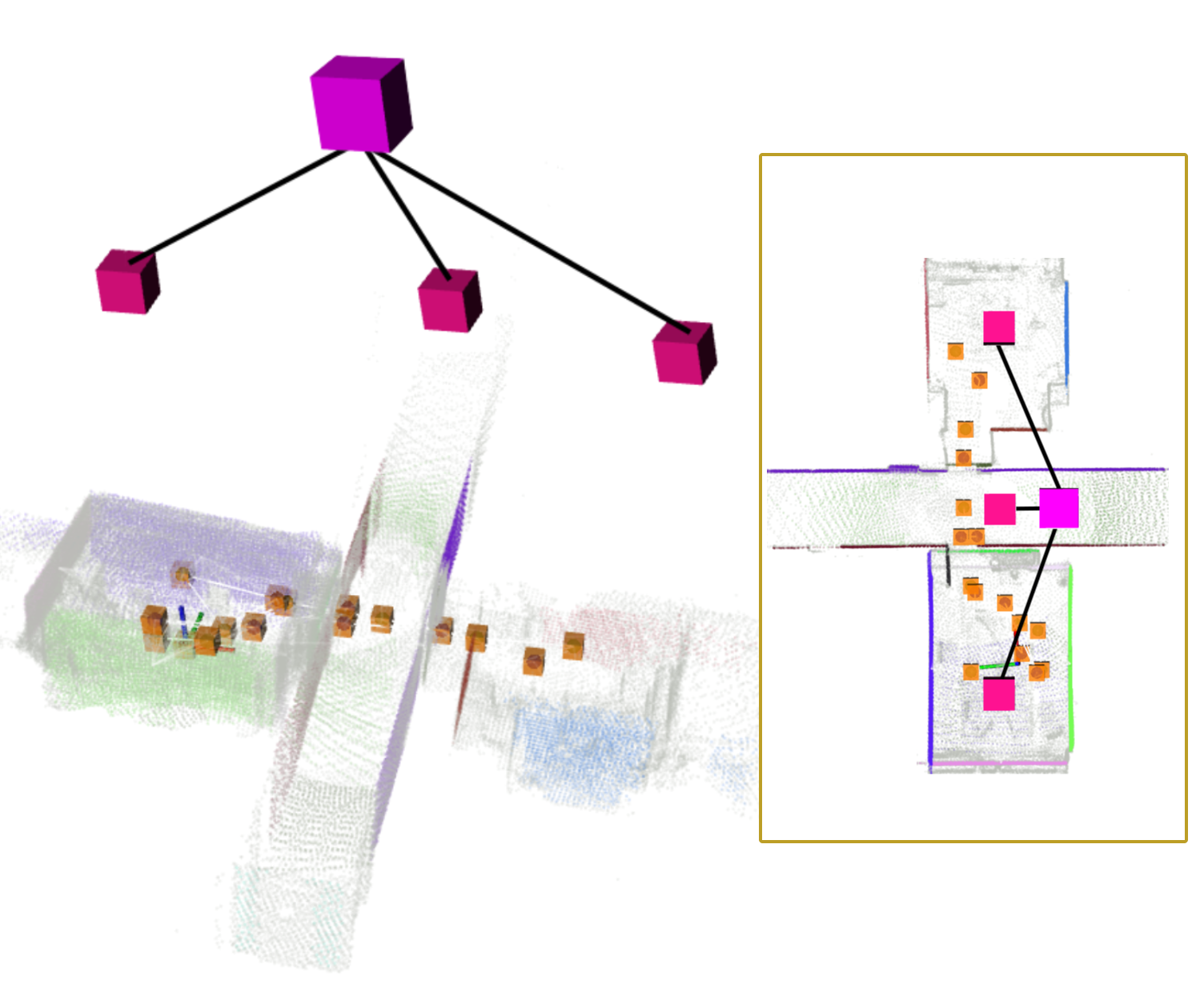}
         \caption{\sgraphs~on \textit{MR01}}
         \label{fig_eval_slam_sgraphs_s06}
     \end{subfigure}
     \begin{subfigure}[t]{0.35\columnwidth}
         \centering
         \includegraphics[width=0.98\columnwidth]{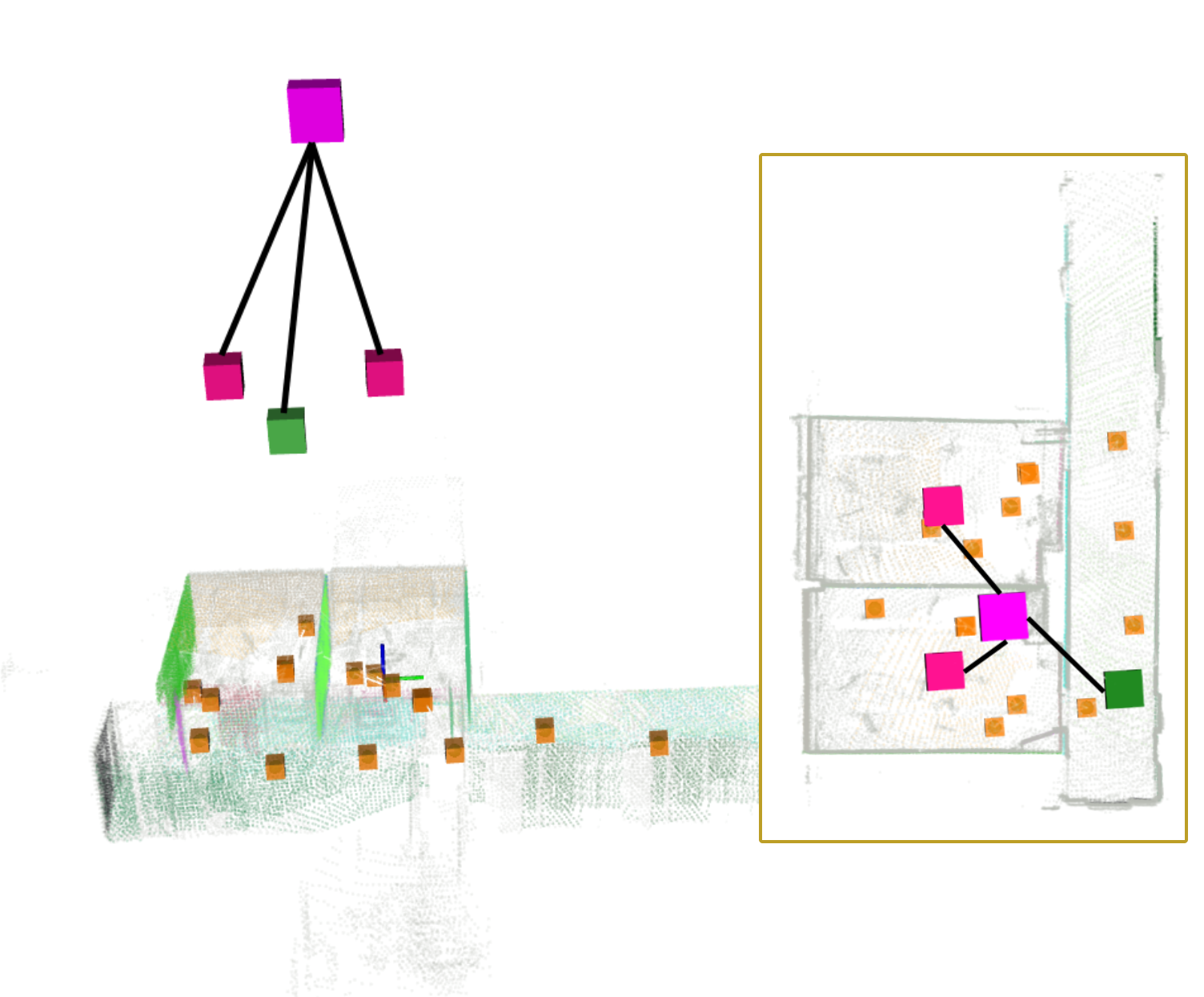}
         \caption{\sgraphs~on \textit{MR02}}
         \label{fig_eval_slam_sgraphs_s07}
     \end{subfigure}
     \begin{subfigure}[t]{0.35\columnwidth}
         \centering
         \includegraphics[width=\columnwidth]{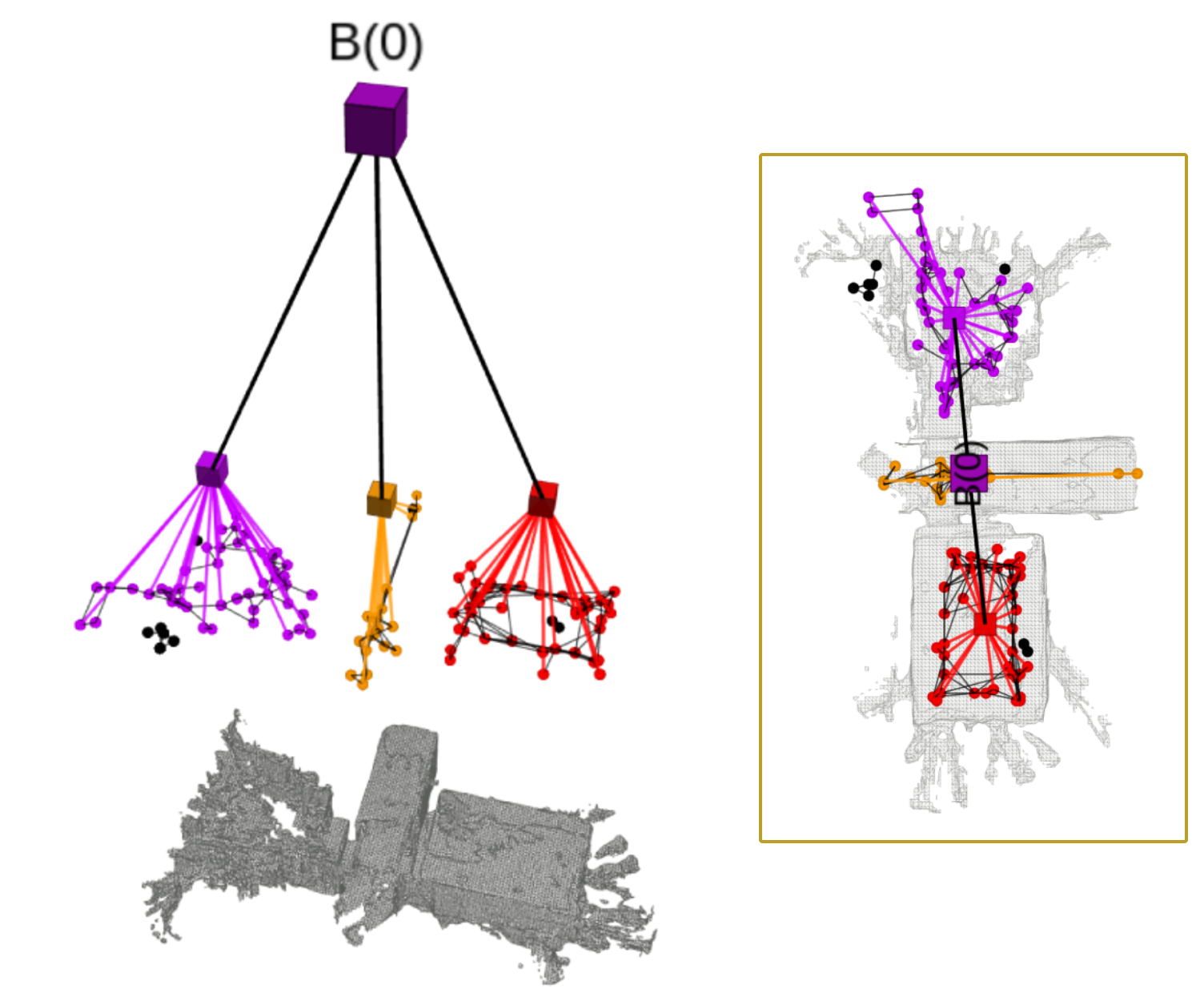}
         \caption{Hydra on \textit{MR01}}
         \label{fig_eval_slam_hydra_s06}
     \end{subfigure}
     \begin{subfigure}[t]{0.35\columnwidth}
         \centering
         \includegraphics[width=\columnwidth]{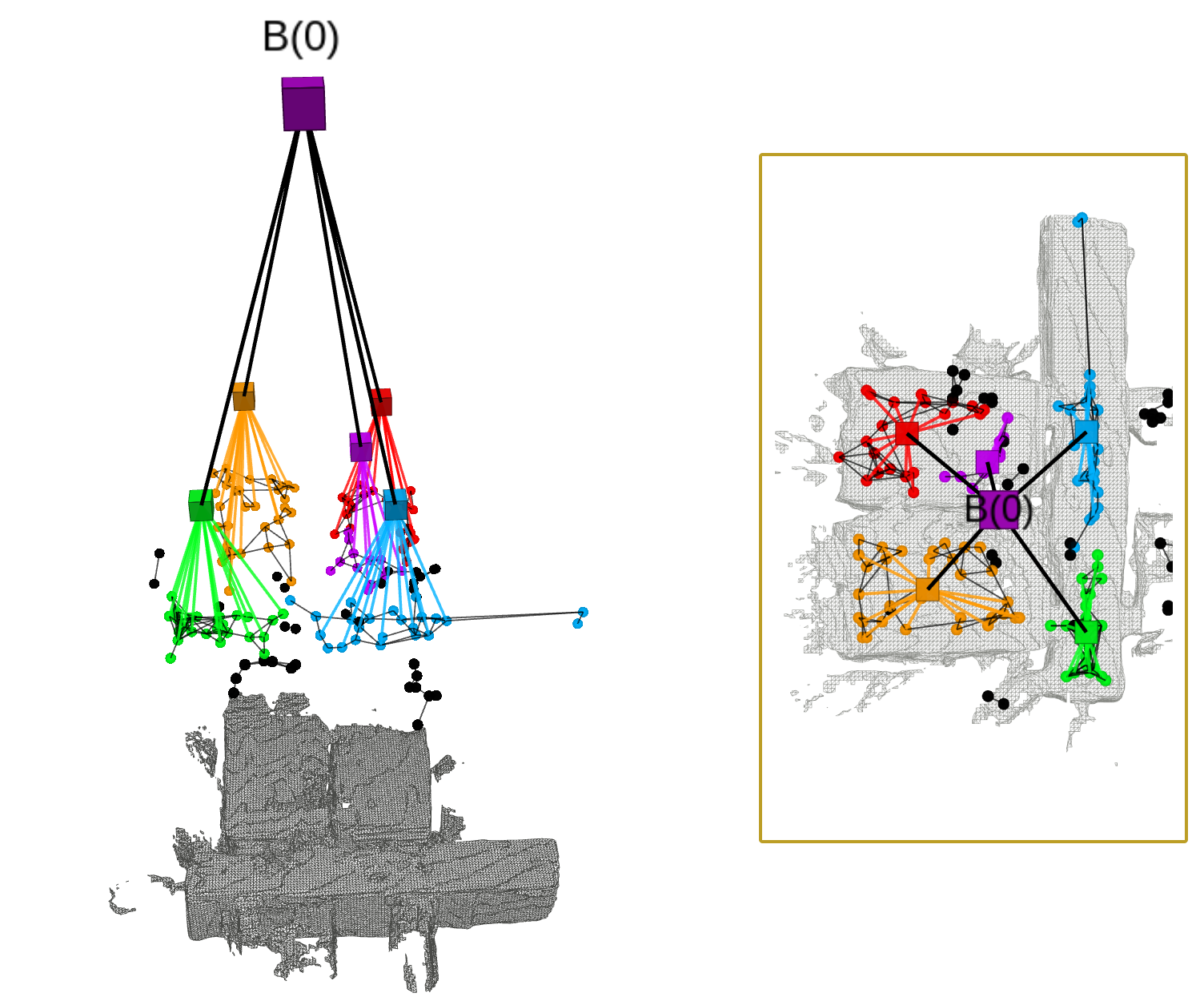}
         \caption{Hydra on \textit{MR02}}
         \label{fig_eval_slam_hydra_s07}
     \end{subfigure}
     \caption{Qualitative comparison of reconstructed scene graphs on multi-room sequences of the \textit{AutoSense} dataset.}
     \label{fig_eval_semantic}
\end{figure}

% Results
Experimental results show that vS-Graphs, despite operating solely on RGB-D input, achieves an accuracy comparable to the LiDAR-based method (\textit{S-Graphs}) in detecting building components and structural elements.
This highlights the effectiveness of our visual feature processing and scene graph generation in capturing structural layouts with high precision.
Notably, vS-Graphs demonstrates higher \textit{recall}, even in larger and more complex environments such as MR03, owing to its visual verification of building components that ensures consistent entity detection across extended scenes.
It should be noted that \enquote{wall} entities are not directly provided in Hydra; therefore, Hydra's performance is assessed based on correct \enquote{room} element counting and recognition.
Additionally, Fig.~\ref{fig_eval_semantic} provides a qualitative comparison of the reconstructed scene graphs generated by vS-Graphs, \textit{S-Graphs}, and Hydra across two representative dataset sequences.
\begin{figure}[!t]
     \centering
     \includegraphics[width=0.9\columnwidth]{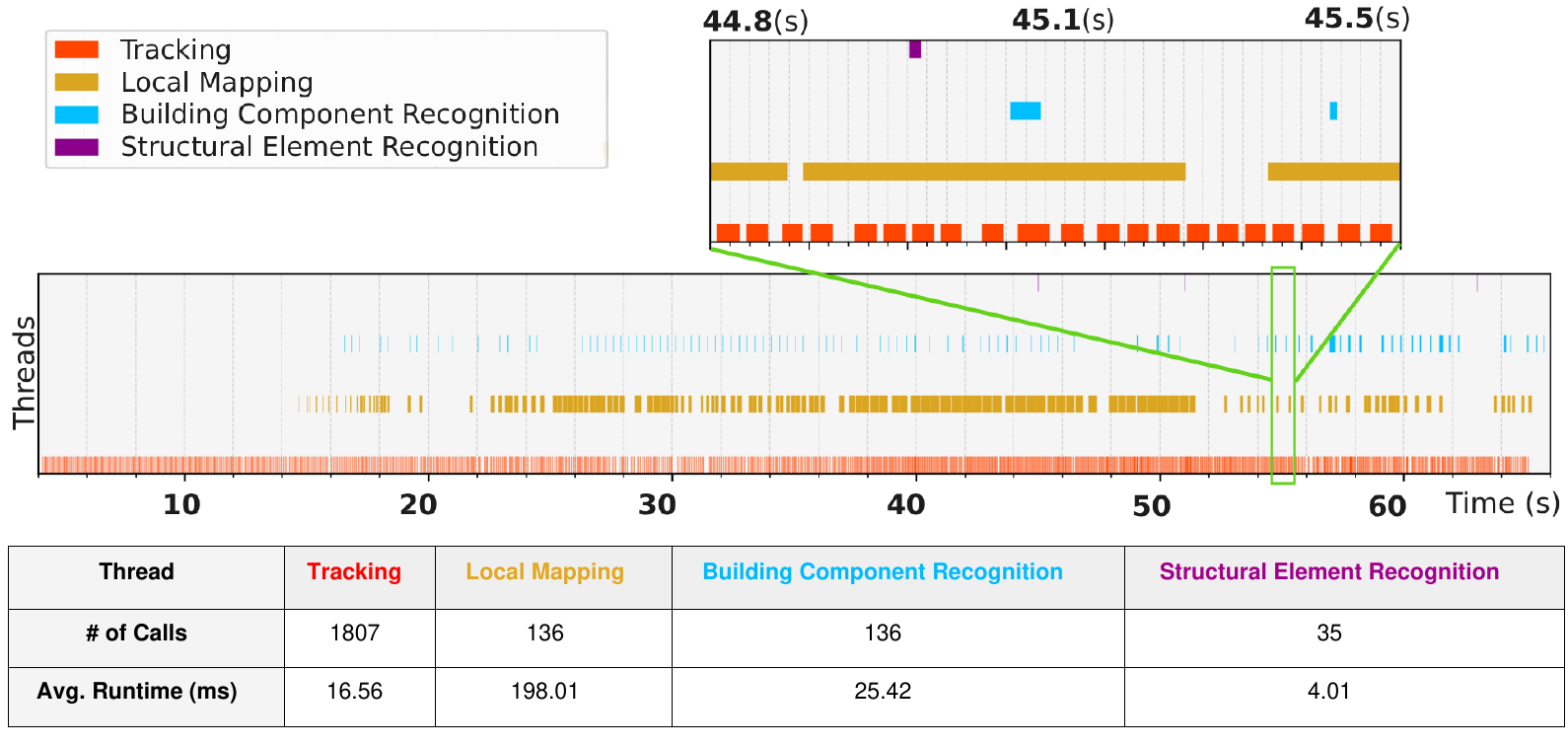}
     \caption{Timeline of thread execution within vS-Graphs while processing a sample dataset instance (sequence \textit{SR01}).}
     \label{fig_eval_profiling}
\end{figure}

\subsubsection{Runtime Analysis}
\label{sec_eval_statistics}

% Realtime
vS-Graphs achieves real-time performance with an average processing rate of $22\pm3~\mathrm{FPS}$, exceeding the $20~\mathrm{FPS}$ threshold considered for real-time operation. % (compared to \orb's $29 \pm 3$ \ac{FPS})
This is accomplished through a multi-threaded architecture, as shown in Fig.~\ref{fig_eval_profiling}.
The \enquote{\textit{Tracking}} thread processes visual features at the frame level, while \enquote{\textit{Local Mapping}} concurrently maps objects and optimizes their positions.
\enquote{\textit{Building Component Recognition}}, running in parallel at the KeyFrame level, identifies potential wall and ground surfaces from the online panoptic segmentation.
The \enquote{\textit{Structural Element Recognition}} runs less frequently and in constant periods (every two seconds) to infer rooms and floors in the map.
Compared to \orb's $29\pm3~\mathrm{FPS}$ on the same hardware and dataset, the slightly reduced frame rate is a reasonable trade-off for vS-Graphs's rich semantic scene understanding.
\section{Conclusions}
\label{sec_conclusions}

% Conclusion
This paper introduced vS-Graphs, a real-time VSLAM framework that reconstructs the robot's operating environment using optimizable hierarchical 3D scene graphs.
It detects building components (wall and ground surfaces), from which structural elements (rooms and floors) are inferred, and incorporates them into hierarchical representations.
Consequently, beyond enhancing map reconstruction by integrating these entities, vS-Graphs offers structured representations of spatial relationships among high-level environment-driven semantic entities.
Experimental results on standard and in-house indoor datasets demonstrated that the framework achieves superior trajectory estimation and mapping performance compared to state-of-the-art VSLAMs, reducing trajectory error by 15.22\% across a wide range of dataset instances.
Other evaluations have shown that the visual features processed by vS-Graphs can identify semantic entities describing the environment's layout with accuracy comparable to that of precise LiDAR-based methods.

% Future Works
Future work includes integrating additional building components (\textit{e.g.,} ceilings, doorways), extending support for complex and concave room layouts via GNN-based approaches, and developing fully semantic-driven loop closure detection, where revisited areas can be identified primarily based on previously detected structural elements and building components.
% Another direction is extending the framework for \textit{multi-floor} setups.
\subsection*{Appendix I. Semantic Augmentation using Fiducial Markers}

% Intro
In vS-Graphs, fiducial markers serve as an \textbf{optional semantic augmentation mechanism}, providing high-level contextual information to the reconstructed map.
Unlike marker-based Visual SLAM methods~\cite{semuco,vsgraphs1,sslam,ucoslam}, where markers play a central role in localization and map construction, in our vS-Graphs system, their role is supportive rather than structural.
The pipeline does not depend on the presence of markers for map accuracy or robustness; instead, they enrich the representation by linking structural elements to semantic labels.
To enable this, we assume that each structural element (\textit{e.g.,} a room) may contain one unique \textbf{ArUco marker}.
This criterion is implemented during the data collection procedures, where unique markers are placed in each free-space zone, such as rooms or corridors, and their associations are stored in a dedicated environment-driven database (\texttt{JSON} file).
For instance, a marker with $\mathrm{marker\_id}=5$ is assigned to \enquote{\textit{corridor-A}}, while marker $\mathrm{marker\_id}=8$ was assigned to \enquote{\textit{office-B}}.
At runtime, if (i) \textbf{the ArUco detector is active} and (ii) \textbf{the corresponding association file (database of marker information) is provided}, the system uses this information to enrich detected structural elements semantically.
Thus, if the pose of a detected marker falls within the area of a recently detected n-wall room, it will be augmented with the semantic label stored in the database.

\begin{figure}[!t]
    \centering
    \includegraphics[width=\textwidth]{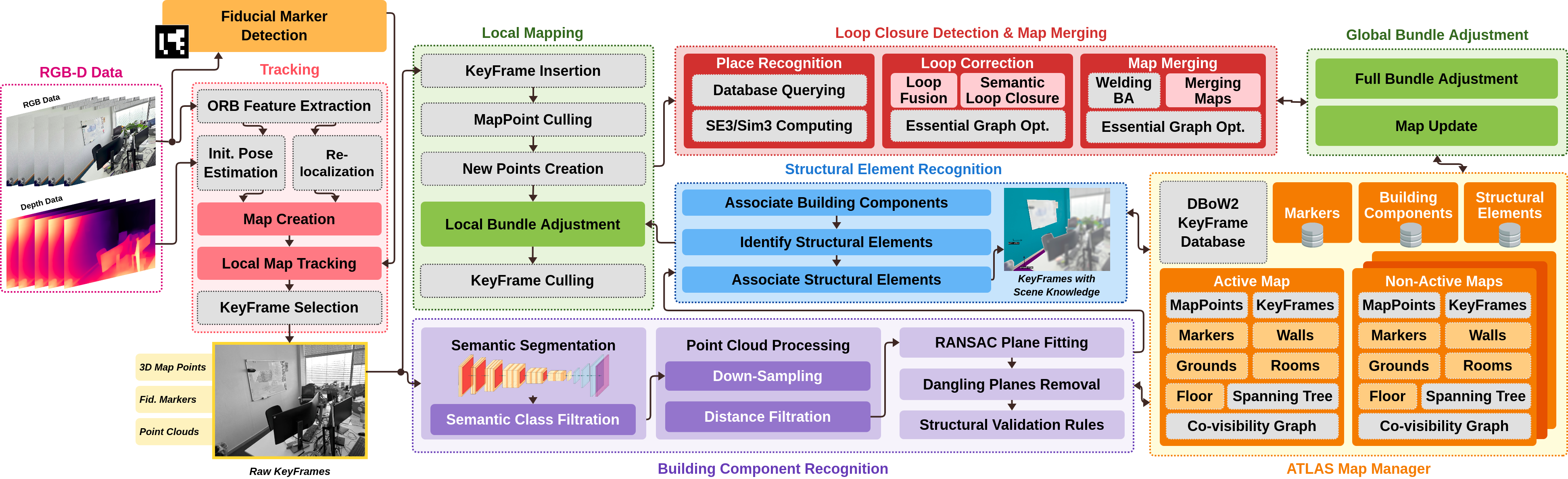}
    \caption{The multi-thread architecture of vS-Graphs with the optional fiducial marker detector module. Modules with dashed borders and a light gray background are inherited directly from the baseline (\textit{i.e.,} ORB-SLAM 3.0), while the remaining components are the contributions of vS-Graphs.}
    \label{fig_flowchart_with_marker}
\end{figure}

% Processing
Fig.~\ref{fig_flowchart_with_marker} shows the overall architecture of vS-Graphs, including the fiducial marker detection and integration module, where recognized markers are stored in the \textit{ATLAS Map Manager}.
In brief, RGB-D data is processed in real time, providing integrated visual and depth information to subsequent modules.
Meanwhile, \enquote{\textit{Fiducial Marker Detection}} (utilizing the ArUco library in this work) operates independently on input frames, identifying markers \textit{only if they are present} in the environment and associating them with the corresponding frames.
Visual features are extracted and tracked across sequential frames in the \enquote{\textit{Tracking}} thread, where pose information is either initialized or refined, depending on the map reconstruction stage, creating a 3D map with tracked features across frames.
KeyFrame selection, a critical step following feature extraction, is performed within the \enquote{\textit{Tracking}} thread by analyzing the visual data.
These KeyFrames contain 3D map points, point clouds, and potentially detected fiducial markers, forming the foundation for subsequent processes.

% Equations
Accordingly, each fiducial marker \(\mathbf{m}_i^K = \{\mathbf{p}_m, \mathbf{\nu}_m\}\) in $G_t$ is constrained by the KeyFrame $K$ observing it, where \(\mathbf{p}_m \in SE(3)\) is the marker's pose and \(\mathbf{\nu}_m\) is its center point.
The mentioned constraint is defined below:
\begin{equation}
    \mathbf{\sigma}_{m}(^G\mathbf{K}, \mathbf{m}_i^K) = \| ^L\mathbf{m}_i \boxplus \leftidx{^G}{\mathbf{K}} \boxminus \leftidx{^G}{\mathbf{m}_{{i_1}}} \|^2_{\mathbf{\Lambda}_{\mathbf{\tilde{m}}_{i}}}
    \label{eq_marker}
\end{equation}
\noindent where $^L\mathbf{m}_{{i}}$ refers to the locally observed marker's pose, $\boxplus$ and $\boxminus$ represent the composition and inverse composition, $\|\dots\|$ is the \textit{Mahalanobis} distance, and $\mathbf{\Lambda}_{\mathbf{\tilde{m}}_{i}}$ is marker's information matrix.

% Marker Association
The marker \(\mathbf{m}_i^{K_j} \in \mathbf{M}\) is associated with a structural element if it lies within the spatial bounds of the detected room/corridor \(\delta^{K_g} \in \mathbf{\Delta}_{\phi}\).
Thus, the Euclidean distance between the center point \(\mathbf{\nu}_m\) of \(\mathbf{m}_i^{K_j}\) and the room/corridor's centroid \(\mathbf{\nu}_s\), expressed as \(d(\mathbf{\nu}_m, \mathbf{\nu}_s) \leq \epsilon_s\), where \(\epsilon_s\) is the proximity threshold.
The supporting condition is that \(\mathbf{\nu}_m\) must be enclosed among all bounding walls \(\mathbf{\Pi}_{wall}\) with the normal vector \(\mathbf{n}\):
\begin{equation}
    \forall \pi_{wall}^{K_i} \in \mathbf{\Pi}_{wall}, \quad (\mathbf{n}_{wall}^{K_i} \cdot (\mathbf{\nu}_m - \mathbf{\nu}_r)) \leq 0
    \label{eq_marker_assoc}
\end{equation}

\begin{figure}[!t]
    \centering
    \includegraphics[width=0.9\columnwidth]{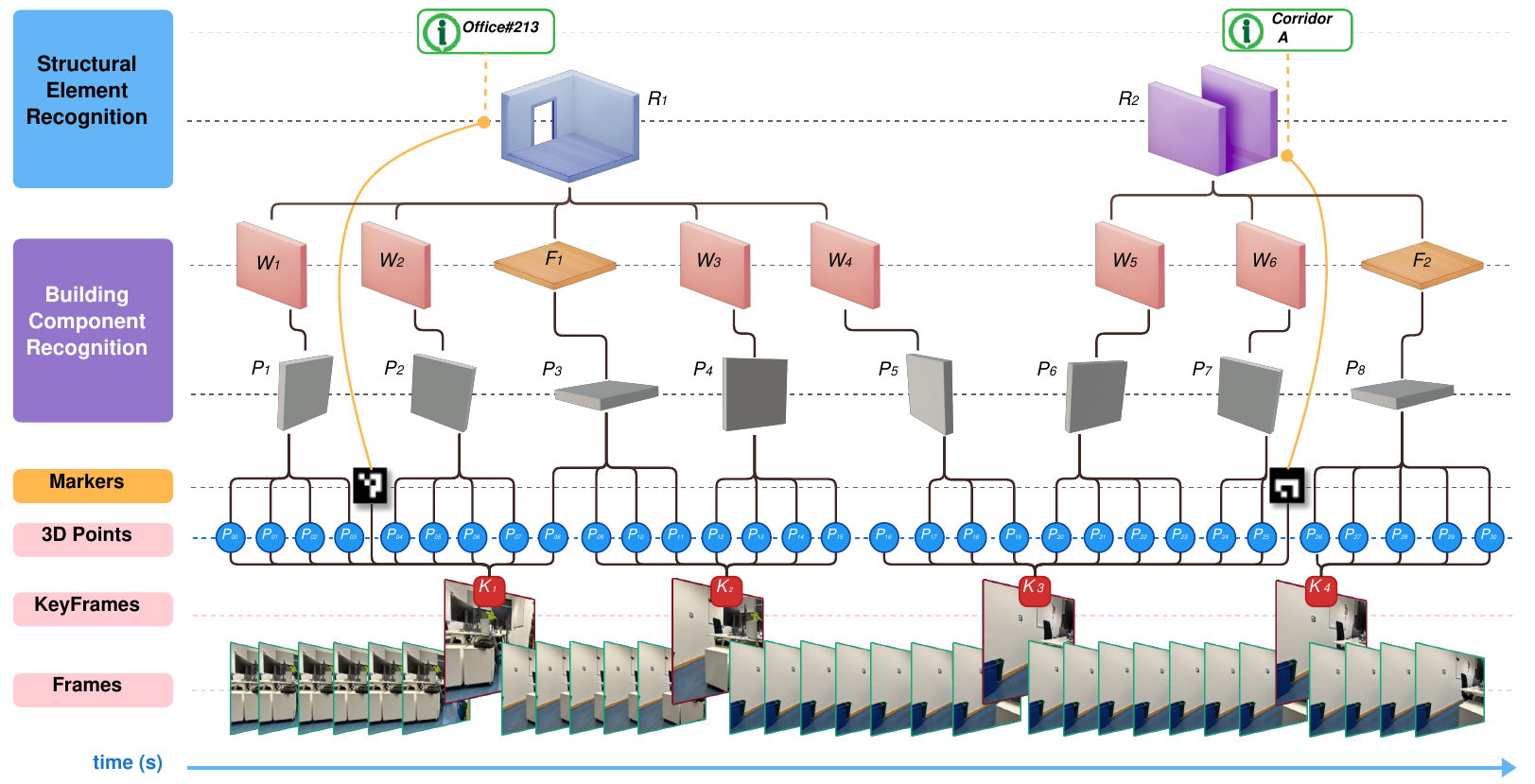}
    \caption{Scene graph structure generated using vS-Graphs, creating a hierarchical representation of the environment.}
    \label{fig_graph_with_marker}
\end{figure}

With this, Fig.~\ref{fig_graph_with_marker} illustrates the extended scene graph structure of the vS-Graphs framework, emphasizing the role of fiducial markers in labeling the detected structural elements.

\clearpage

\subsection*{Appendix II. Mapping Performance Relative to Baseline}
\begin{table}[!h]
    \centering
    \caption{Root Mean Square Error (RMSE) values for ORB-SLAM 3.0 and vS-Graphs across different sequences of the SMapper dataset (over eight iterations). The results indicate that vS-Graphs generally achieves lower RMSE values.}
    \begin{tabular}{l|l|c|c|c|c|c|c|c|c|c|c}
        \toprule
            \textbf{Method} & \textbf{Sequence} & \multicolumn{8}{c}{\textbf{Iteration Index}} \\
        \cmidrule(lr){3-10}
        & & 1 & 2 & 3 & 4 & 5 & 6 & 7 & 8 \\
        \midrule
            \multirow{6}{*}{\rotatebox{45}{vS-Graphs}} & SR01 & 0.2598 & 0.3102 & 0.2608 & 0.3454 & 0.3409 & 0.3122 & 0.2891 & 0.3246 \\
            & SR02  & 0.2333 & 0.2165 & 0.2301 & 0.2494 & 0.2499 & 0.2414 & 0.2564 & 0.2853 \\
            & SR03 & 0.2749 & 0.5544 & 0.2426 & 0.3509 & 0.2837 & 0.3801 & 0.2725 & 0.2920 \\
            & MR01  & 4.3774 & 5.5194 & 6.2763 & 4.7143 & 6.8155 & 4.4640 & 4.8322 & 4.5823 \\
            & MR02 & 1.1153 & 1.1836 & 0.9430 & 0.9017 & 0.9210 & 1.0799 & 0.9575 & 0.7648 \\
            & MR03  & 1.2933 & 0.2978 & 0.3472 & 0.3137 & 1.1864 & 0.2894 & 0.4122 & 0.2721 \\
        \midrule
            \multirow{6}{*}{\rotatebox{45}{ORB-SLAM 3.0}} & SR01 & 0.2772 & 0.4435 & 0.3509 & 0.3964 & 0.2753 & 0.3006 & 0.3995 & 0.3602 \\
            & SR02  & 0.3111 & 0.2781 & 0.2862 & 0.2668 & 0.3000 & 0.2955 & 0.2408 & 0.2787 \\
            & SR03 & 0.3451 & 0.3380 & 0.3279 & 0.2931 & 0.3054 & 0.3728 & 0.3076 & 0.3435 \\
            & MR01  & 5.1266 & 5.0866 & 5.6484 & 5.6531 & 4.7557 & 6.2066 & 5.2847 & 5.2242 \\
            & MR02 & 1.1828 & 1.1521 & 0.9291 & 0.9648 & 0.9379 & 1.0269 & 1.2719 & 1.0016 \\
            & MR03  & 0.2869 & 2.5426 & 1.5765 & 2.3271 & 2.0381 & 0.2725 & 0.6926 & 0.4120 \\
        \bottomrule
    \end{tabular}
\end{table}

\begin{table}[!h]
    \centering
    \caption{Number of map points generated by ORB-SLAM 3.0 and vS-Graphs on the SMapper dataset (over eight iterations). The measurements show that vS-Graphs produces fewer points than ORB-SLAM 3.0, while improving mapping accuracy.}
    \begin{tabular}{l|l|c|c|c|c|c|c|c|c|c|c}
        \toprule
            \textbf{Method} & \textbf{Sequence} & \multicolumn{8}{c}{\textbf{Iteration Index}} \\
        \cmidrule(lr){3-10}
        & & 1 & 2 & 3 & 4 & 5 & 6 & 7 & 8 \\
        \midrule
            \multirow{6}{*}{\rotatebox{45}{vS-Graphs}} & SR01 & 6759  & 6704  & 6744  & 6741  & 6875  & 6804  & 6780  & 6842  \\
            & SR02  & 7304  & 7120  & 7420  & 7457  & 7421  & 7212  & 7382  & 7160  \\
            & SR03 & 11764 & 11591 & 11937 & 12054 & 12156 & 11844 & 11471 & 11475 \\
            & MR01  & 20607 & 20682 & 19692 & 20714 & 21042 & 20274 & 19433 & 20546 \\
            & MR02 & 16838 & 16622 & 16479 & 16563 & 17065 & 17200 & 16905 & 17274 \\
            & MR03  & 48364 & 46237 & 48250 & 47850 & 47149 & 49008 & 48442 & 45224 \\
        \midrule
            \multirow{6}{*}{\rotatebox{45}{ORB-SLAM 3.0}} & SR01 & 6931  & 7085  & 7057  & 6903  & 7012  & 6894  & 7063  & 7130  \\
            & SR02  & 7639  & 7378  & 7548  & 7712  & 7540  & 7590  & 7363  & 7377  \\
            & SR03 & 13140 & 12631 & 12818 & 12903 & 13021 & 13187 & 12613 & 12990 \\
            & MR01  & 22564 & 22685 & 21916 & 22688 & 22552 & 22759 & 22701 & 22506 \\
            & MR02 & 18816 & 18196 & 17899 & 18547 & 17722 & 18492 & 18117 & 17643 \\
            & MR03  & 54797 & 55342 & 56531 & 55499 & 56056 & 55853 & 56187 & 54545 \\
        \bottomrule
    \end{tabular}
\end{table}

\clearpage

\subsection*{Appendix III. Generated Scene Graphs (Qualitative Analysis)}

\begin{figure}[!ht]
     \centering
     \begin{subfigure}[t]{0.32\textwidth}
         \centering
         \includegraphics[width=\textwidth]{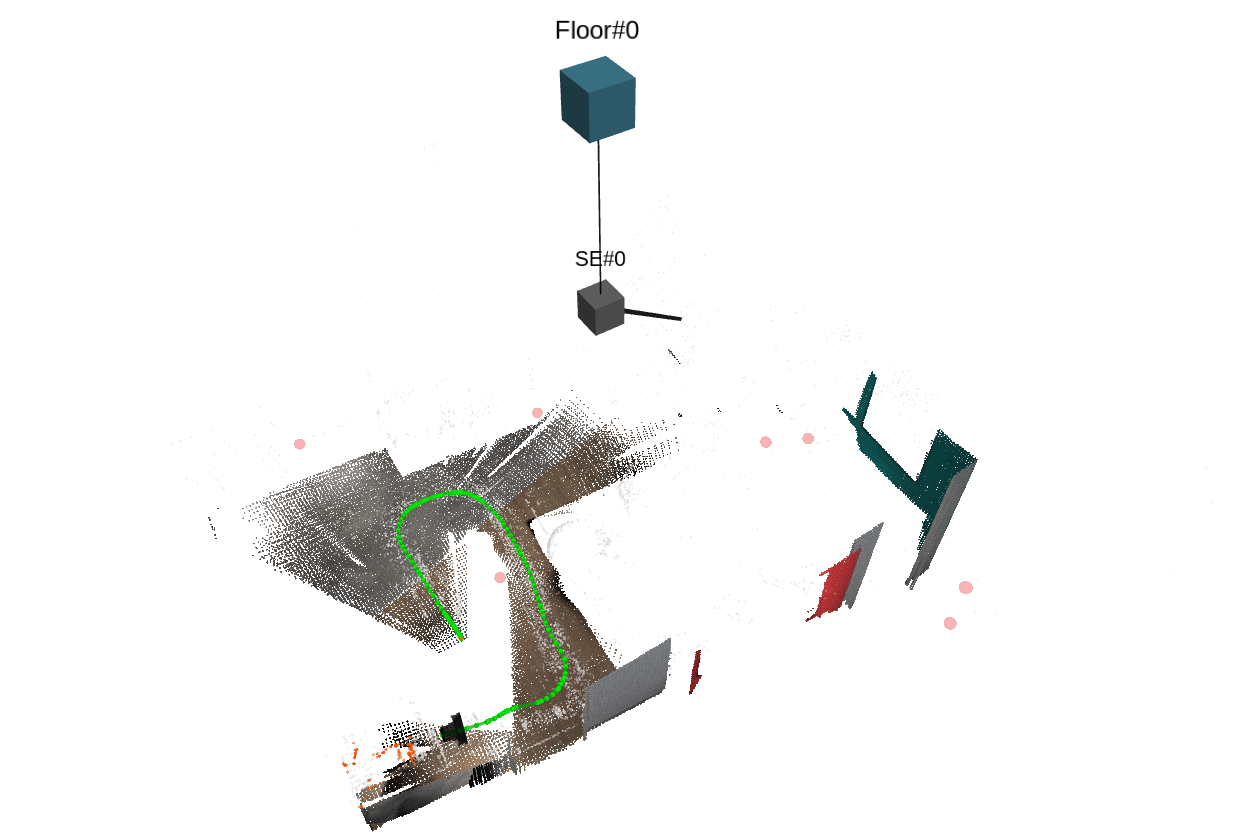}
         \caption{ICL - deer-gr.}
     \end{subfigure}
     % \hspace{1mm}
     \begin{subfigure}[t]{0.32\textwidth}
         \centering
         \includegraphics[width=\textwidth]{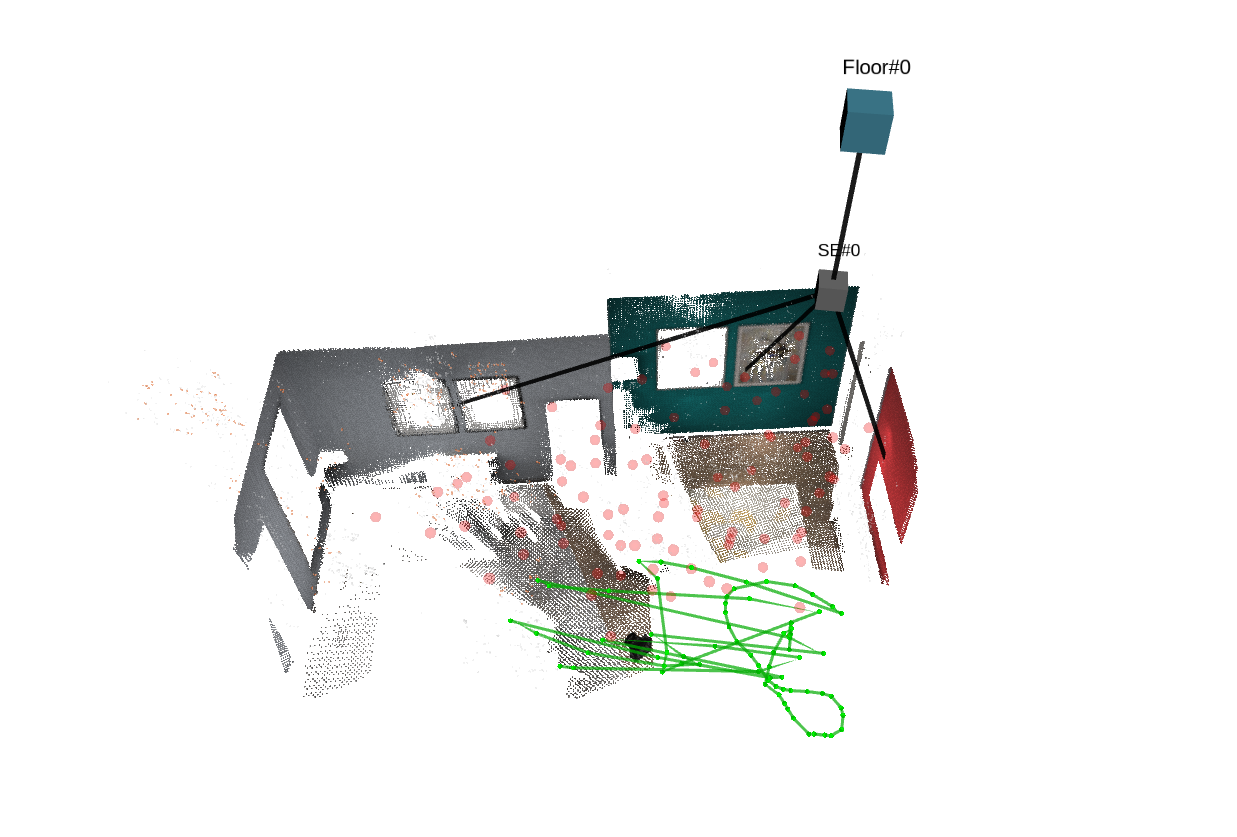}
         \caption{ICL - deer-mavf.}
     \end{subfigure}
     % \hspace{1mm}
     \begin{subfigure}[t]{0.32\textwidth}
         \centering
         \includegraphics[width=\textwidth]{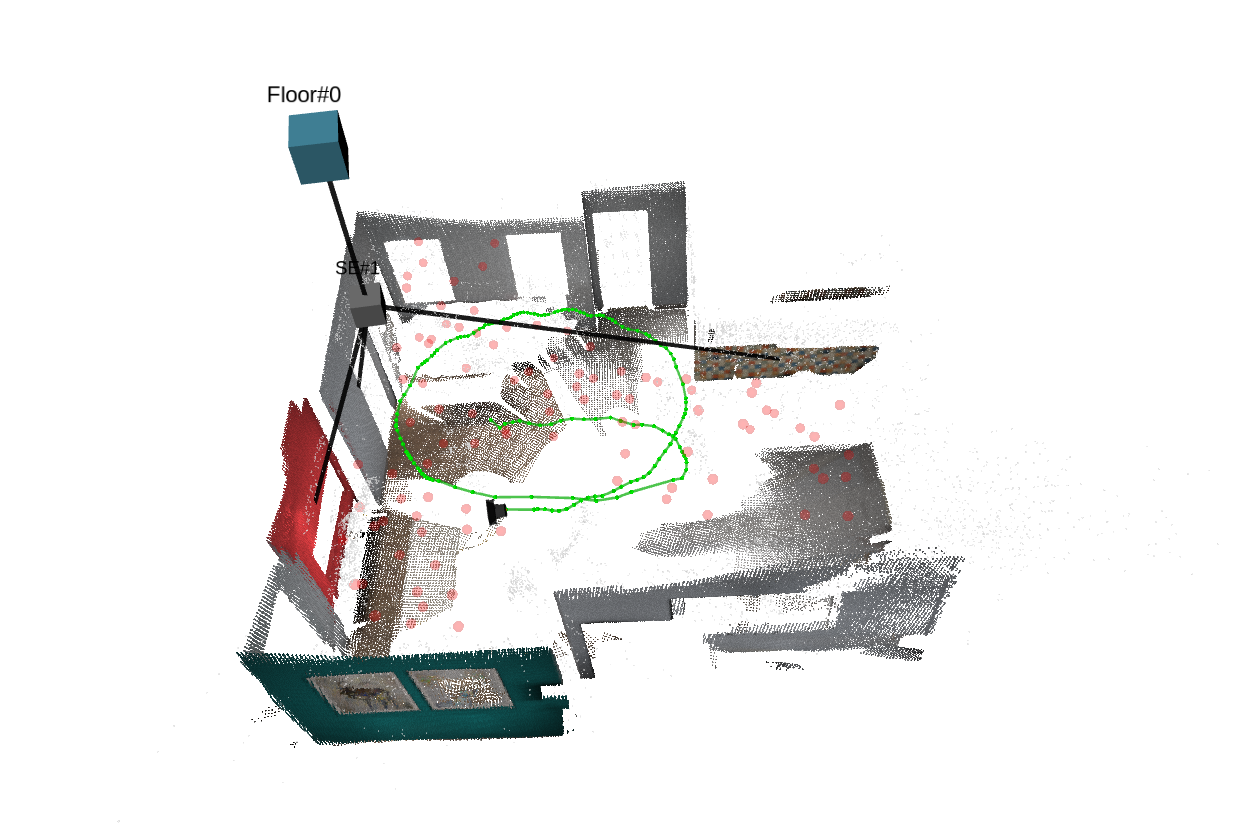}
         \caption{ICL - deer-r.}
     \end{subfigure}
     % ---------------
     \begin{subfigure}[t]{0.32\textwidth}
         \centering
         \includegraphics[width=\textwidth]{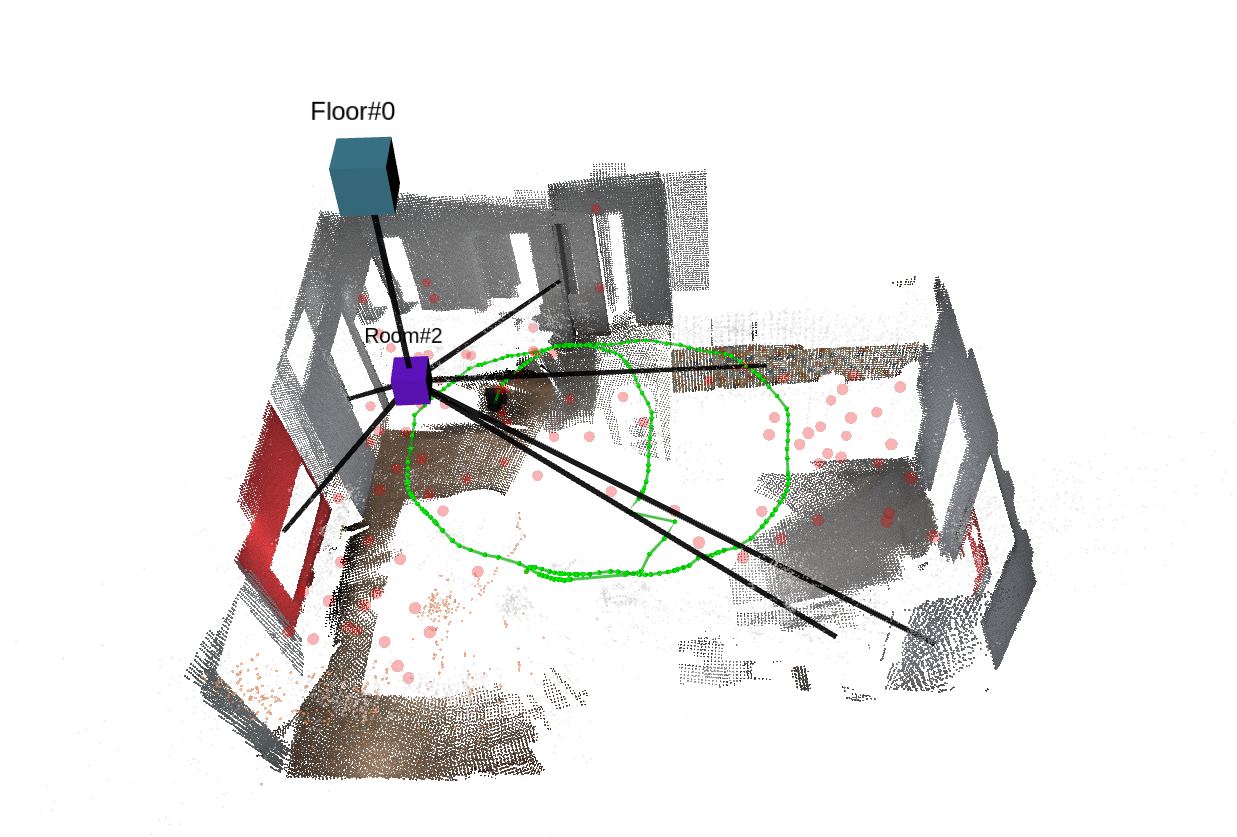}
         \caption{ICL - deer-w.}
     \end{subfigure}
     \begin{subfigure}[t]{0.32\textwidth}
         \centering
         \includegraphics[width=\textwidth]{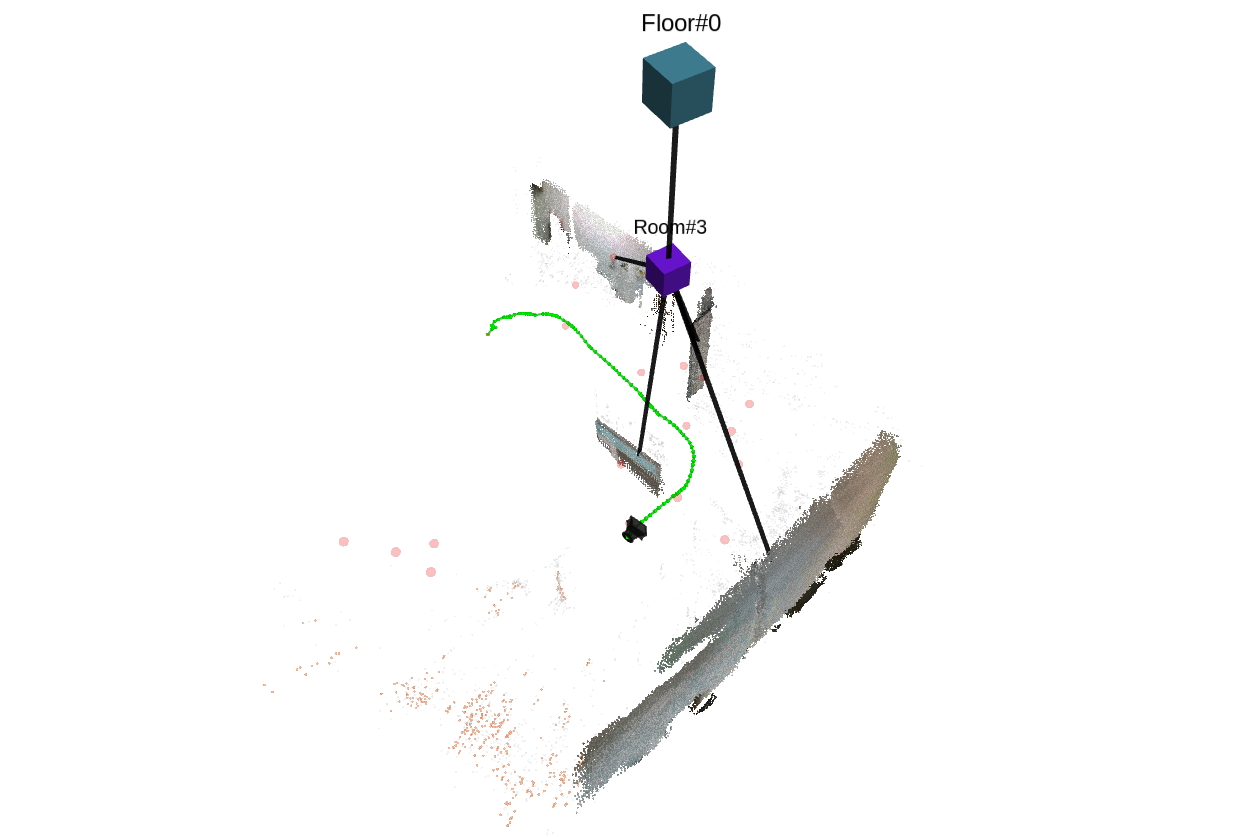}
         \caption{OpenLoris - office-1-2.}
     \end{subfigure}
     \begin{subfigure}[t]{0.32\textwidth}
         \centering
         \includegraphics[width=\textwidth]{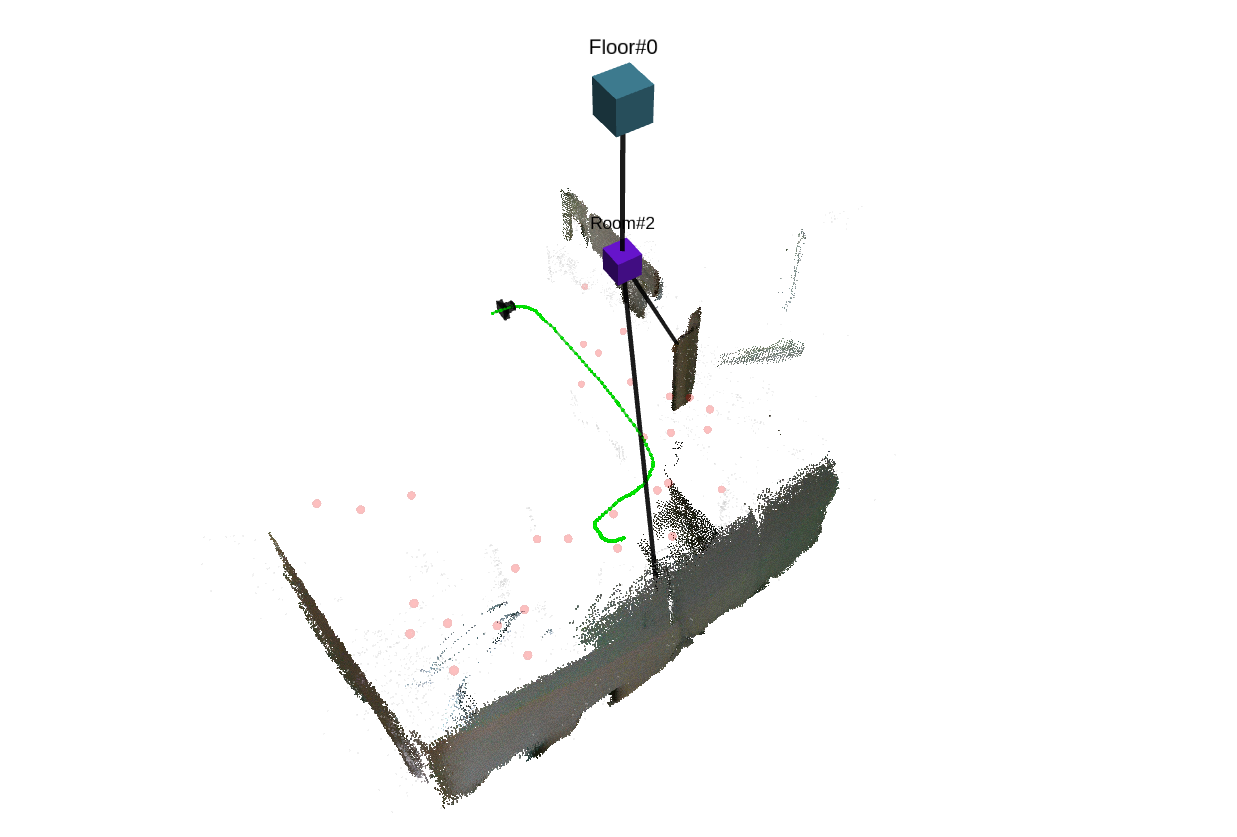}
         \caption{OpenLoris - office-1-4.}
     \end{subfigure}
     \begin{subfigure}[t]{0.32\textwidth}
         \centering
         \includegraphics[width=\textwidth]{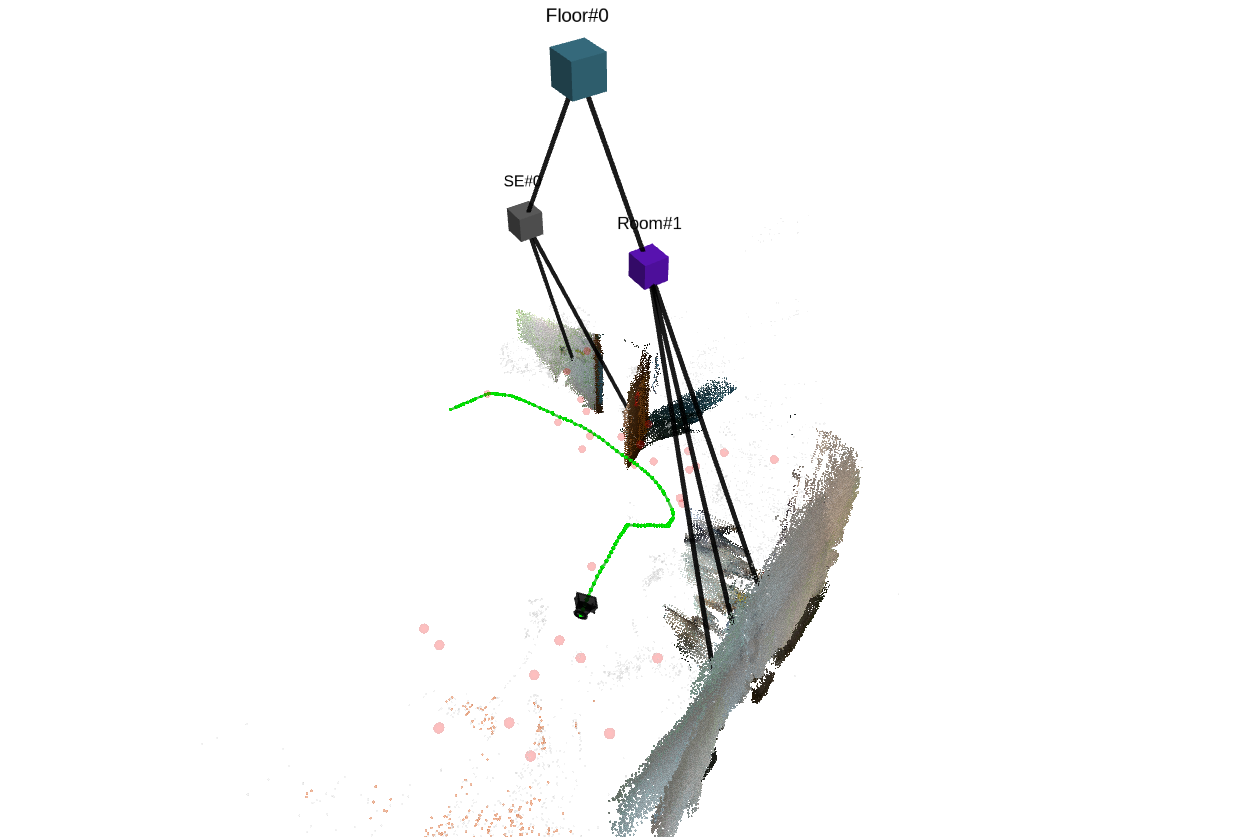}
         \caption{OpenLoris - office-1-7.}
     \end{subfigure}
     \begin{subfigure}[t]{0.32\textwidth}
         \centering
         \includegraphics[width=\textwidth]{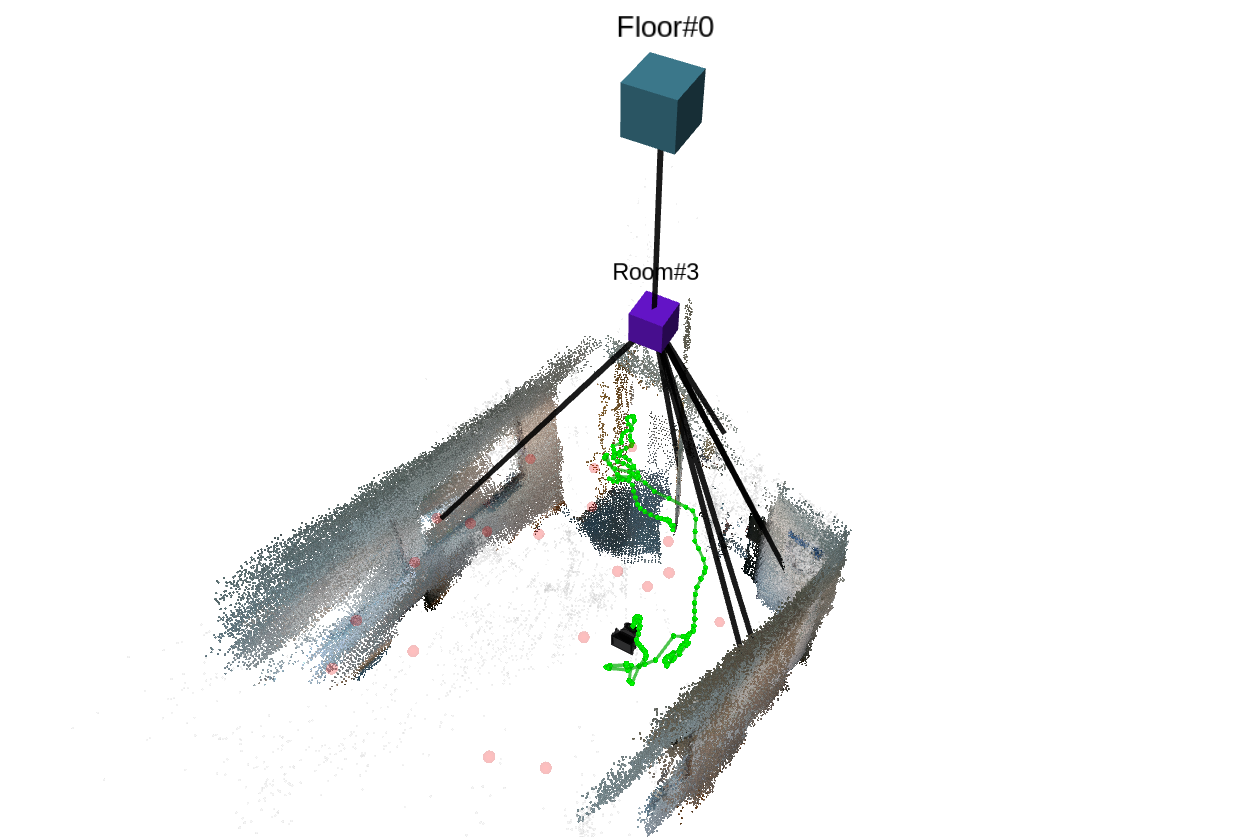}
         \caption{SMapper - SR02.}
     \end{subfigure}
     \begin{subfigure}[t]{0.32\textwidth}
         \centering
         \includegraphics[width=\textwidth]{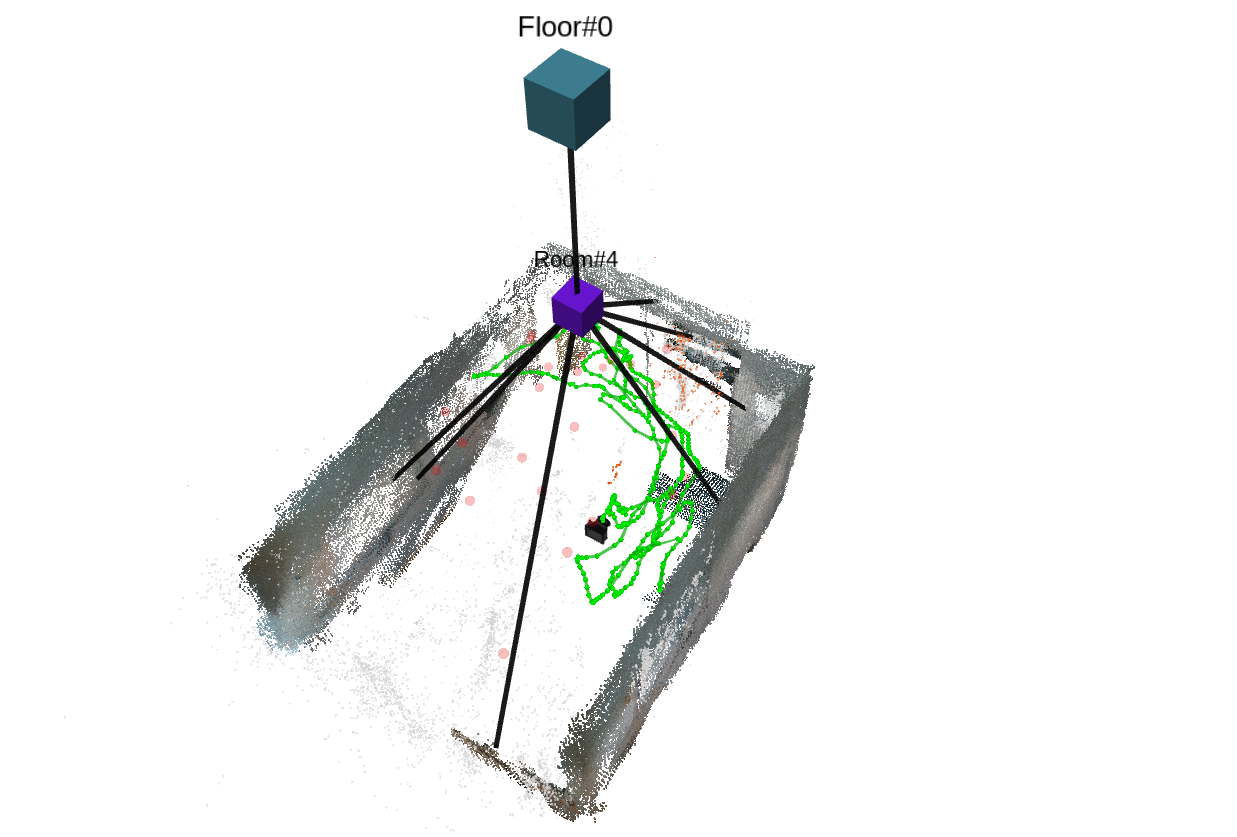}
         \caption{SMapper - SR03.}
     \end{subfigure}
     \begin{subfigure}[t]{0.32\textwidth}
         \centering
         \includegraphics[width=\textwidth]{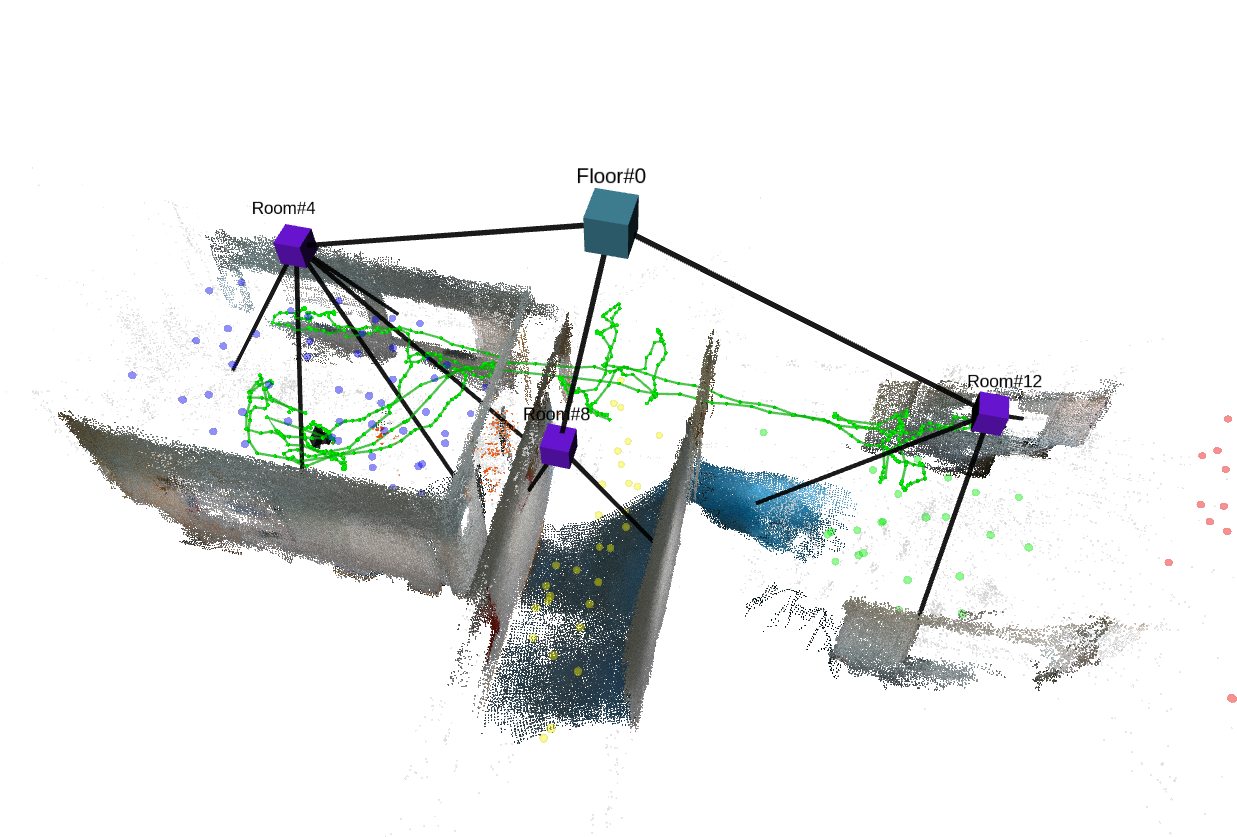}
         \caption{SMapper - MR01.}
     \end{subfigure}
     \begin{subfigure}[t]{0.32\textwidth}
         \centering
         \includegraphics[width=\textwidth]{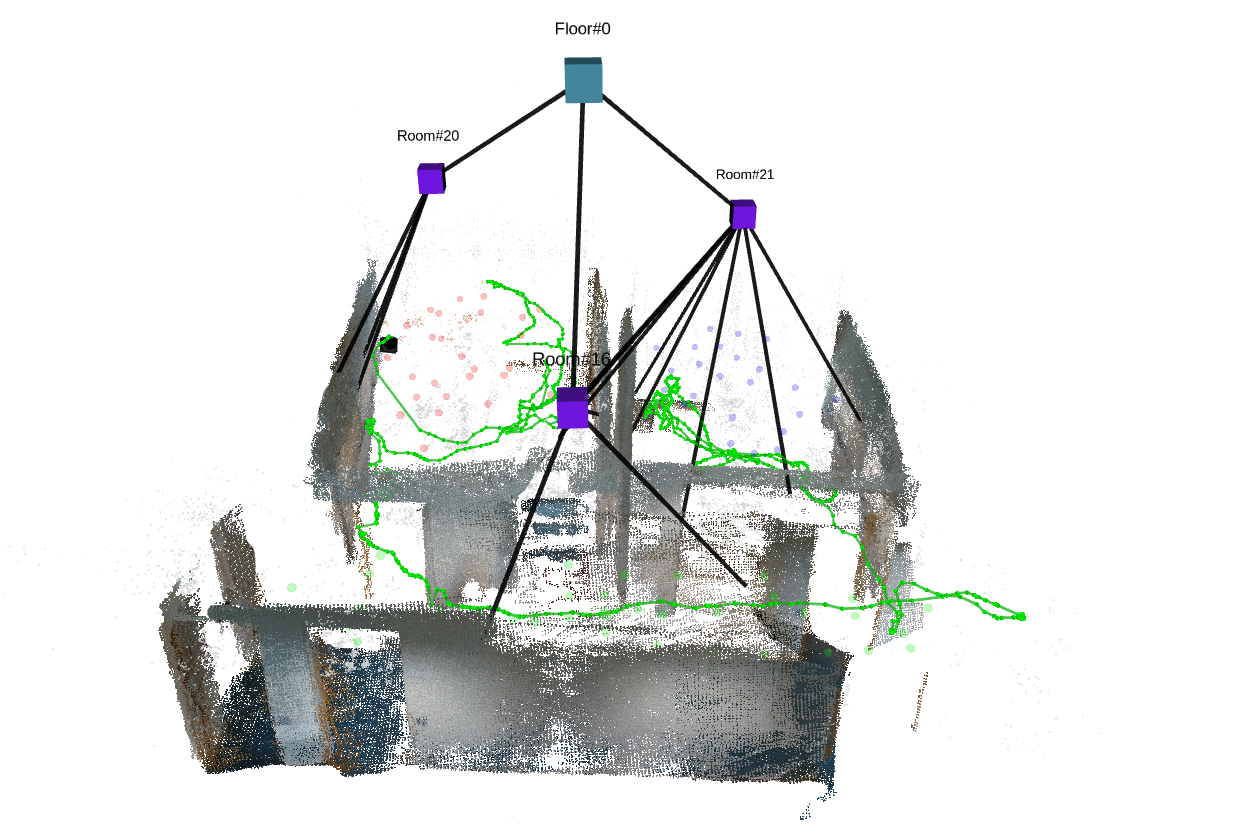}
         \caption{SMapper - MR02.}
     \end{subfigure}
     \begin{subfigure}[t]{0.32\textwidth}
         \centering
         \includegraphics[width=\textwidth]{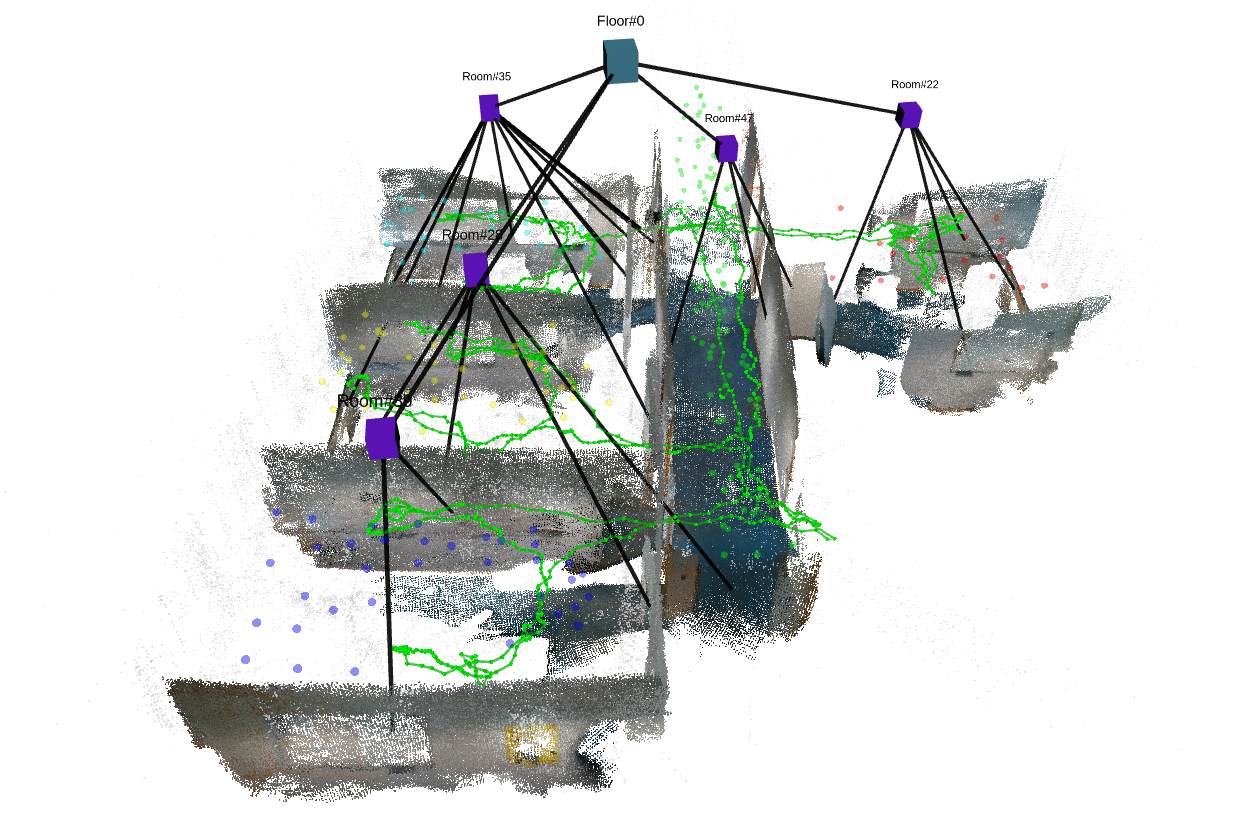}
         \caption{SMapper - MR03.}
     \end{subfigure}
     \caption{Scene graph examples generated by vS-Graphs across different environments, demonstrating consistent hierarchical semantic representation of structural components.}
\end{figure}

% \addtolength{\textheight}{-12cm}

\newpage

% References (reads from the bibliography file
\bibliographystyle{IEEEtran}
\bibliography{root}

\end{document}